\documentclass{article}

\usepackage[utf8]{inputenc} % allow utf-8 input
\usepackage[T1]{fontenc}    % use 8-bit T1 fonts
\usepackage{hyperref}       % hyperlinks
\usepackage{url}            % simple URL typesetting
\usepackage{booktabs}       % professional-quality tables
\usepackage{amsfonts}       % blackboard math symbols
\usepackage{nicefrac}       % compact symbols for 1/2, etc.
\usepackage{graphicx}
\usepackage{fancyhdr} % https://www.overleaf.com/learn/latex/Headers_and_footers
\newcommand{\hlf}{\textstyle \frac{1}{2} \displaystyle}
\newcommand{\reals}{{\rm I\kern-.17em R}}
\def\Reals{{\hbox{$\it I\hskip-3.6pt R$}}}

\def\Reals{{\hbox{$\it I\hskip-3.6pt R$}}}

\newcommand{\ie}{{\em i.e.}}

\newcommand{\etal}{{\em et al.}}

\newcommand{\calL}{{\cal L}}

\newcommand{\keiko}{\stackrel{\triangle}{=}}

\newcommand{\implies}{{\Rightarrow}}

\title{Learning Shared Kernel Models: the Shared Kernel EM algorithm}

\author{
  Graham W. Pulford, BandGapAI, France\\
}

\date{20 May 2020}
\begin{document}
\maketitle

\begin{abstract}
Expectation maximisation (EM) is an unsupervised learning method for estimating the parameters of a finite mixture distribution. It works by introducing  ``hidden'' or ``latent'' variables via Baum's auxiliary function $Q(\Theta,\Theta_0)$ that allow the joint data likelihood to be expressed as a product of simple factors. The E-step of EM computes the expectation over the hidden variables. The M-step refines the parameter estimates $\Theta_0$ by maximising $Q(\Theta,\Theta_0)$ over the parameter space $\Theta$. The relevance of EM has increased since the introduction of the variational lower bound (VLB): the VLB differs from Baum's auxiliary function only by the entropy of the PDF of the latent variables $Z$. We first present a rederivation of the standard EM algorithm using data association ideas from the field of multiple target tracking, using $K$-valued scalar data association hypotheses ($K\geq 2$) rather than the usual binary indicator vectors. The same method is then applied to a little known but much more general type of supervised EM algorithm for shared kernel models, related to probabilistic radial basis function networks. We address a number of shortcomings in the derivations that have been published previously in this area. In particular, we give theoretically rigorous derivations of (i) the complete data likelihood; (ii) Baum's auxiliary function (the E-step) and (iii) the maximisation (M-step) in the case of Gaussian shared kernel models. The subsequent algorithm, called shared kernel EM (SKEM), is then applied to a digit recognition problem using a novel 7-segment digit representation. Variants of the algorithm that use different numbers of features and different EM algorithm dimensions are compared in terms of mean accuracy and mean IoU. A simplified classifier is proposed that decomposes the joint data PDF as a product of lower order PDFs over non-overlapping subsets of variables. The effect of different numbers of assumed mixture components $K$ is also investigated. High-level source code for the data generation and SKEM algorithm is provided.
\end{abstract}

\noindent
{\bf Keywords}: EM algorithm, supervised EM algorithm, expectation maximization, mixture model, shared kernel model, shared kernel EM, SKEM algorithm, shared mixture model, shared mixture classifier, class-conditioned model, Baum's auxiliary function, hidden variable, latent variable, probabilistic radial basis function, data association, 7-segment digit model, supervised learning.

\tableofcontents
\newpage

\section{Introduction}
Parameter estimation for finite mixture models is inherently ill conditioned: methods that seek directly to maximise the likelihood function exhibit poor convergence. Moreover, the maximum likelihood objective function is intractable since the number of terms is ${\rm O}(K^N)$ where $K$ is the number of mixture components and $N$ is the number of data samples. This is clearly exponential in the number of samples. The Expectation Maximisation (EM) algorithm introduced in 1970 \cite{Baum2}, is the main method for obtaining mixture parameter estimates in a numerically tractable way. In essence, it introduces a set of auxiliary variables called ``hidden'' or ``latent'' variables, via which the joint PDF is most easily expressed. The hidden variables provide the categorical information that assigns to each data sample a component in the mixture model \cite{Titterington}. Since these variables are unknown, an iterative approach is adopted wherein the hidden variables are ``averaged out'' of the joint density at each iteration. The parameters are re-estimated by optimising the resulting averaged density and the process is repeated until convergence.

It is well known that the EM algorithm is related to the K-means clustering algorithm, which is less general than EM because it assumes that all clusters have the same shape, i.e., covariance matrix. Both  are unsupervised learning techniques that can be applied to multi-dimensional data sets. The EM algorithm automatically clusters the data into a number of component densities whose weights $\pi_k$ are the {\em prior} probabilities of pertaining to the classes or categories represented by the components densities.

Although not widely recognised, the EM algorithm is an iterative {\em data association technique}, with the hidden variables playing the role of {\em data association hypotheses}. The degree of membership of each data sample to the mixture components is quantified by a set of {\em posterior} weights $w_{nk}$ that play the role of {\em association probabilities}, as they are known in the multiple target tracking literature. Thus there is a link between hidden variables in the EM algorithm and target tracking methods like probabilistic data association (PDA) \cite{Barshalom1} and joint PDA \cite{Fortmann2} in terms of their discrete probabilistic structure. One contribution of this paper is to show how Baum's auxiliary function, which is the basis of the E-step in the EM algorithm, is constructed via $K$-valued data association hypotheses rater than the usual binary 0/1 variables in the statistics literature. This is presented in section \ref{emalg}.

The conventional EM algorithm estimates the parameters $\Theta$ in a mixture model of the form
\begin{equation}\label{icdl}
{\rm p}(x|\Theta)=\sum_{j=1}^{L}\Pr(c=j)\,{\rm p}(x|c=j,\theta_j)
\end{equation}
where $c$ is the class, $\Pr(c=j)$ is the prior class probability and there are $L$ classes. This is naturally just a reexpression of the law of total probability where the events $\{c=j\}$ are mutually exclusive and exhaustive. The terms ${\rm p}(x|c=j,\theta_j)$ are class-conditioned PDFs. If the latter are Gaussian, then ${\rm p}(x|\Theta)$ is a Gaussian mixture density or Gaussian mixture model.

In contrast, the main concern of this paper is parameter estimation for {\em shared kernel} models (SKMs). This type of model arises when we allow each class-conditioned density to originate from a model in which the kernels, or component densities, are shared between the classes, with different classes having different weights. We can represent this as
\begin{equation}
{\rm p}(x|c=j,\Theta)=\sum_{k=1}^{K}\pi_{kj}{\rm p}(x;\theta_k)
\end{equation}
where $\pi_{kj}$ is the prior probability that the data sample $x$ belongs to class $j$ and is associated with mixture component $k$. The total probability mass for each of the $L$ classes is now distributed among the $K$ components, so the $\pi_{kj}$ sum to 1 over $k$. This flexibile representation allows the data PDF of each class to represented as a mixture density, rather than just a component of a mixture density. If the kernels are Gaussian, then ${\rm p}(x;\theta_k)={\rm N}\{x;\mu_k,P_k\}$ and each class-conditioned density is a Gaussian mixture. As implied by the notation, all class-conditioned densities use the same set of kernels, or, in the Gaussian case, all have the same mean vectors $\mu_k$ and covariance matrices $P_k$.

The shared kernel model was present in a rudimentary form in a paper by Luttrell \cite{Luttrell}, who coined the terms ``multiple overlapping mixture distributions'' and ``partitioned mixture distributions.'' The idea was formalised in \cite{Jarrad}, where it was referred to as a ``shared mixture distribution.'' The actual terminology of ``shared kernel model'' is due to Titsias and Likas \cite{Titsias}, who investigated links to feedforward neural network structures employing radial basis functions (RBFs). Subsequently the same authors renamed their method to ``class conditional mixture density with constrained component sharing'' \cite{Titsias2003}, although this paper contains no algorithm derivations. Loosely speaking, the ``EM algorithm for probabilistic RBF training'' in \cite{Titsias} and the ``non-discriminative EM'' algorithm in \cite{Jarrad} are ``equivalent'' to the algorithm presented in this paper. Both preceding papers, however, fail to provide a correct derivation of Baum's auxiliary function  $Q(\Theta,\Theta_0)$, and of the subsequent modified EM algorithm required for SKMs with Gaussian kernels. The EM algorithms in these papers will not work properly on Gaussian shared kernel models if implemented as described.

To avoid confusion, it should be pointed out that there are also variants of the EM algorithm that constrain either the structure of the mixture covariance matrices or share the covariance matrix between different model components (e.g., \cite{Dharanipragada}): these are merely simplifications of the standard EM algorithm and should not be confused with the algorithm developed here.

A more general version of the EM algorithm can also be developed \cite{Neal99,Bishop2009} via the so-called variational lower bound (VLB), which calls for a functional or variational maximisation in the E-step. An understanding of the original EM algorithm is a pre-requisite for understanding variational Bayesian inference since the VLB differs from Baum's auxiliary function only by the entropy of the PDF of the latent variables $Z$. A thorough understanding of the role of hidden variables in the EM algorithm will be beneficial to the AI community due to the close connection between hidden variables in EM and latent variables in the VLB. These aspects are explored further in a companion publication \cite{Pulford2020b}.

The main contribution of the present paper is to provide theoretically rigorous derivations of (i) the complete data likelihood; (ii) Baum's auxiliary function (the E-step) and (iii) the subsequent maximisation (M-step) in the case of Gaussian shared kernel models. The derived algorithm is called {\em shared kernel expectation maximisation} (SKEM), which is the subject of section \ref{skem}. In contrast to standard EM, the SKEM algorithm, which works with class-conditioned mixtures, requires class labels for its training. SKEM is therefore a supervised learning method.

The derivation of the SKEM algorithm relies on the machinery of probabilistic data association (introduced in section \ref{emalg}). The E-step requires explicit enumeration of all {\em sequences} of data association hypotheses in order to compute the expectation over the hidden variables. To achieve this end, we introduce class-conditioned set notation that correctly accounts for arbitrary partitions of the data set into class-indexed sets. This essential part of the calculation was glossed over in  \cite{Jarrad} and \cite{Titsias}, resulting in expressions for the complete data likelihood and Baum's $Q$ function that superficially look correct but contain errors.  The derivation in the general case is clarified in a low-order worked example in section \ref{wex}.

Again, unlike previous work on shared kernel learning, we include extensive numerical simulations on a digit recognition problem in section \ref{NS}. This uses a novel image generation method based on ``7-segment'' displays to produce 28 x 28 pixel greyscale images of synthetic digits that can be used to train the SKEM algorithm. These digits contain the same number of pixels as those in the MNIST database \cite{LeCun2}. The data generation process is controlled by a small number of adjustable parameters and is quite flexible.

In section \ref{algsimpl}, ten different implementations of the SKEM algorithm for this problem are given, depending on the choice of features and the dimensions of the variables used in the SKEM loop. Digit classification is on the basis of a maximum likelihood criterion. Performance metrics such as mean accuracy and mean intersection over union (IoU) are obtained on 500 Monte Carlo runs (see section \ref{results}). Examples of 10-class confusion matrices are also given. The use of random initialisation allows the sensitivity of the SKEM algorithm to be compared for various classifier implementations. An interesting by-product of the numerical testing is that the performance of higher dimensional implementations can be enhanced by decomposing the joint likelihood into a product of lower dimensional likelihoods obtained from non-overlapping subsets of features. A brief investigation into the effect of the choice of the number of mixture components $K$ appears in section \ref{Kvar}. Matlab$^{\rm TM}$ source code is given for the data generation, feature extraction and SKEM algorithm in section \ref{src}.

To improve readability, we provide some examples here of class-conditioned mixtures obtained using the SKEM algorithm (see figures on next page). The two sets of plots show the 15 2-D projections (for a single class) of a 6-D shared kernel mixture with 10 mixture components. Line thickness is indicative of the weight of mixture component (see Table \ref{Tab7} in section \ref{plots} for an explanation).

\begin{figure}[p]
\center
   \includegraphics[width=14cm]{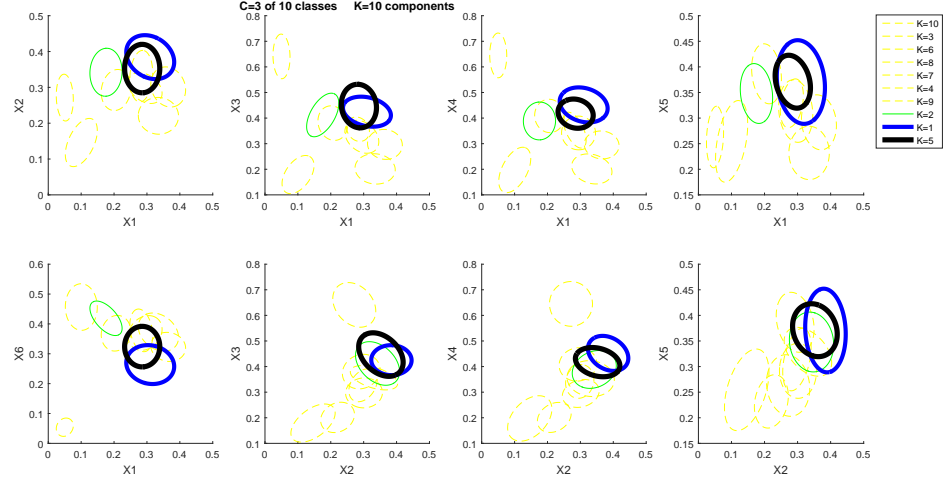}
\caption{First eight 2-D projections of a 6-D shared kernel mixture obtained with SKEM algorithm using 6 PCA features. Line thickness indicates weight of mixture component, with insignificant components in dashed yellow.}
\label{fig6d1}
\end{figure}

\begin{figure}[p]
\center
   \includegraphics[width=14cm]{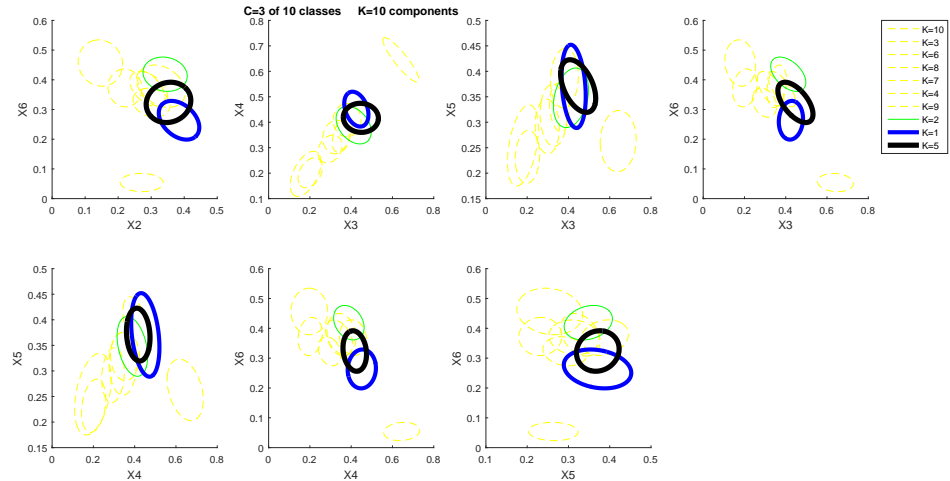}
\caption{Remaining seven 2-D projections of same 6-D shared kernel mixture in Fig. \ref{fig6d1}.  Mixture weight encoded by line thickness. Plots for other classes with same shared kernel appear in section \ref{plots}.}
\label{fig6d2}
\end{figure}

\section{Expectation Maximisation}\label{emalg}
\subsection{Learning the Parameters of a Finite Mixture Density}\label{EMalg}
The following treatment of the EM algorithm adopts notation from chapter 9 of \cite{Bishop2009}, which is more in line with the notation typically used in the AI literature than the notation found in the statistics literature.
\begin{itemize}
    \item Incomplete data $X=\{x_1,\ldots,x_N\},~x_i\in\Reals^n$, also called unlabelled or uncategorised data, depending on the problem context.
    \item Missing data $Z=\{z_1,\ldots,z_N\}$, also called hidden or latent variables.
    \item Complete data $Y=(X,Z)$, the union of the incomplete and missing data.
    \item Mixture parameters $\Theta$. In the case of a $K$-component multivariate Gaussian mixture - the set of parameters of the mixture consisting of (scalar) weights $\pi_i$, mean vectors $\mu_i$,  covariance matrices $P_i$.
\end{itemize}
A basic assumption of the EM algorithm is that each data point $x_i$ is from one and only one component $j$ of a finite mixture distribution. The missing information refers to the component of the mixture pertaining to each observation. In the case of a finite mixture distribution, each element of the missing data vector is an indicator (or label) of the associated mixture component, one per data point. In the case of a Gaussian mixture, the PDF takes the form:
\[
{\rm p}(x)=\sum_{k=1}^K \pi_k {\rm N}(x; \mu_k,P_k)
\]
where ${\rm N}(x; \mu,P)$ is a multivariate Gaussian PDF in the variable $x$, with mean vector $\mu$ and covariance matrix $P$, \ie,
\begin{equation}\label{gaussian}
{\rm N}(x; \mu,P)=(2\pi)^{-n_x/2}({\rm det}P)^{-\hlf}\exp\{-\hlf(x-\mu)^TP^{-1}(x-\mu)\}
\end{equation}
where $n_x$ is the dimension of $x$.
The mixture weights satisfy $\pi_k\geq 0$ $\forall k$ and
\[
\sum_{k=1}^K \pi_k=1,
\]
from which it follows that $\pi_k={\rm Pr}(x\leftrightarrow\mbox{component k})$ in the sense of an association probability. The set of parameters of the Gaussian mixture is denoted $\Theta=\{\pi_k, \mu_k,P_k\}_{k=1}^K$. As previously mentioned, the event $z_i=j$, $j\in\{1,\ldots,K\}$ signifies that observation $x_i$ is attributed to mixture component $j$, or, in other words, the conditional distribution of $x_i$ given $z_j$ is
\[
x_i|z_i\sim {\rm N}(x;\mu_j,P_j)
\]
The standard criterion for estimating the parameters $\{\Theta\}$ of a mixture distribution from incomplete observations $X$ using the EM algorithm is maximum likelihood (ML), although other criteria, such as maximum a posteriori can also be used. In the case of ML, the parameters are obtained by optimising the objective function:
\begin{equation}\label{em1}
    \hat{\Theta}=\arg\max_{\Theta} {\rm p}(X|\Theta)
\end{equation}
Even in the case of independent observations (white noise), the likelihood function for a mixture with unlabelled or uncategorised observations (\ie, when the labels $z_i$ are unavailable) takes the form
\begin{equation}\label{pxtheta}
{\rm p}(X|\Theta)=\prod_{i=1}^N {\rm p}(x_i|\Theta)=\prod_{i=1}^N
\sum_{k=1}^K \pi_k {\rm N}(x_i; \mu_k,P_k)
\end{equation}
Taking the log of the likelihood affords only  limited simplification, {\em viz.}
\begin{equation}\label{logpxtheta}
\log{\rm p}(X|\Theta)=\sum_{i=1}^N
\log\left\{ \sum_{k=1}^K \pi_k {\rm N}(x_i; \mu_k,P_k)\right\}
\end{equation}
No further simplification of the log likelihood is possible in general, and we are left with a numerically ill-conditioned multi-dimensional optimisation problem to obtain the mixture parameters. The ill conditioning arises from the possibility that one or more of the mixture means can be exactly equal to one of the data points $x_i$. In this case the conditional Gaussian PDF behaves as ${\rm det}(P_j)^{-\hlf}$, which tends to infinity when the determinant of the covariance becomes small. Illustrations of this phenomenon appear in \cite{Bishop2009}.

The preceding arguments motivate the expectation-maximisation (EM) algorithm, first introduced by Baum \etal~\cite{Baum2,Dempster}. Instead of the direct ML optimisation in (\ref{em1}), Baum et al. introduced the {\em auxiliary function} $Q(\Theta,\Theta_0)$, which is defined as
\begin{eqnarray}\label{baum}
    Q(\Theta,\Theta_0)&=&{\rm E}_Z[\log {\rm p}(X,Z|\Theta) | X,\Theta_0]\\
    &=&\int \log({\rm p}(X,Z|\Theta))\,{\rm p}(Z | X,\Theta_0)\,dZ\nonumber
\end{eqnarray}
where the expectation is with respect to the joint density of the hidden variables $Z$ given the incomplete data $X$ and some nominal parameter estimates $\Theta_0$. Since $\varphi(\cdot)=\log(\cdot)$ is a convex function, one can apply Jensen's inequality $\varphi({\rm E}[U])\leq{\rm E}[\varphi(U)]$ for any random variable $U$, to show that
\[
\Theta_1=\arg\max_{\Theta} Q(\Theta,\Theta_0) \Rightarrow {\rm p}(X|\Theta_1) \geq {\rm p}(X|\Theta_0)
\]
This makes it feasible to replace the maximisation (\ref{em1}) by an iterative procedure that seeks at each stage (or ``pass'') $p$ the maximum of Baum's auxiliary function. The resulting EM algorithm consists of a loop with two steps as follows
\begin{enumerate}
    \item Expectation (E-step) :
    \[
Q(\Theta,\Theta_{p-1})={\rm E}_Z[\log {\rm p}(X,Z|\Theta) | X,\Theta_{p-1}]
    \]
    \item Maximisation (M-step) :
 \[
\Theta_p=\arg\max_{\Theta} Q(\Theta,\Theta_{p-1})
 \]
\end{enumerate}
Note that the algorithm requires multiple passes through the entire data set $X$ to continually refine the parameter estimates $\Theta_p$. There is no guarantee that the algorithm will converge to the ML estimator since there can be multiple stationary points in the likelihood function, and, depending on the initial parameter estimates, the EM algorithm may converge to any of these.

The utility of the EM algorithm rests on the suppositions that (i) the E-step is explicit and (ii) the M-step is easier to calculate than direct maximisation of the likelihood function. For mixture distributions from the exponential family, which includes finite Gaussian mixtures, this is the case. In such cases, the conditioning on the hidden variables allows the joint data likelihood (for independent observations) to be expressed as
\[
{\rm p}(Y|\Theta)=\prod_{i=1}^N{\rm p}(x_i,z_i|\Theta)=\prod_{i=1}^N{\rm p}(x_i|z_i,\Theta)
\Pr(z_i|\Theta)=\prod_{i=1}^N \pi_{z_i}{\rm N}(x_i;\mu_{z_i},P_{z_i})
\]
This at once makes clear the rationale for forming the complete data likelihood, namely, that the latter quantity is expressible as a product of Gaussian PDFs and mixture weights. Taking the logarithm of this results in a manageable expression
\[
\log\,{\rm p}(Y|\Theta)=\sum_{i=1}^{N}\log\,\pi_{z_i}+\sum_{i=1}^{N}\log\,{\rm N}(x_i;\mu_{z_i},P_{z_i})
\]
 whose expectation with respect to the hidden variables can be explicitly obtained (for the E-step). Contrast this with the incomplete data likelihood in (\ref{pxtheta}), the number of terms in which increases exponentially with the length of the data sequence.

In order to derive the EM algorithm, we must now calculate Baum's auxiliary function and then maximise it with respect to the parameters. We present a detailed derivation of the E-step below and then specialise it to the Gaussian mixture case.

Starting from (\ref{baum}), and noting that the hidden variables are discrete, we have
\begin{eqnarray}
    Q(\Theta,\Theta_0)&=&{\rm E}_Z[\log {\rm p}(X,Z|\Theta) | X,\Theta_0]\label{baum2a}\\
    &=&\sum_{z_1=1}^{K}\cdots\sum_{z_N=1}^{K}\log{\rm p}(X,Z|\Theta)\Pr(Z|X,\Theta_0)\label{baum2b}
\end{eqnarray}
Factorisation of the joint density of the complete data is possible due to the assumed independence of the data samples. The right hand side of (\ref{baum2b}) therefore decomposes into a sum of log-PDF terms each of which involves a product of probabilities of the hidden variables. To simplify the resulting calculations, we define
\begin{eqnarray}
g(z_i,x_j) &=& \log{\rm p}(z_i,x_j|\Theta)\\
h(z_i|x_j) &=& \Pr(z_i|x_j,\Theta_0)\nonumber
\end{eqnarray}
With this shorthand, Baum's auxiliary function becomes
\begin{equation}\label{baum3}
Q(\Theta,\Theta_0)=\sum_{z_1=1}^{K}\cdots\sum_{z_N=1}^{K}\sum_{i=1}^{N}g(z_i,x_i)
\prod_{j=1}^N h(z_j|x_j)
\end{equation}
Upon expanding the inner sum over $i=1:N$ and rearranging we have
\begin{eqnarray*}
Q(\Theta,\Theta_0) &=& \sum_{z_1=1}^{K}g(z_1,x_1)h(z_1|x_1)\sum_{z_2=1}^{K}\cdots\sum_{z_N=1}^{K}\prod_{j\neq 1} h(z_j|x_j)+\cdots\\
 &&+~\sum_{z_N=1}^{K}g(z_N,x_N)h(z_N|x_N)\sum_{z_1=1}^{K}\cdots\sum_{z_{N-1}=1}^{K}\prod_{j\neq N} h(z_j|x_j)
\end{eqnarray*}
which consists of $N$ $N$-fold sum-products. Now consider the $(N-1)$-fold sum in the last part of the above expression:
\[
\sum_{z_1=1}^{K}\cdots\sum_{z_{N-1}=1}^{K}\prod_{j=1}^{N-1} h(z_j|x_j)=
\sum_{z_1=1}^{K}\cdots\sum_{z_{N-2}=1}^{K}\prod_{j=1}^{N-2}h(z_j|x_j)
\sum_{z_{N-1}=1}^{K}h(z_{N-1}|x_{N-1})
\]
The definition of the hidden variables means that for any one of the $z_i$
\[
\sum_{z_i=1}^{K}h(z_i|x_i)=\sum_{k=1}^{K}h(z=k|x_i)=\sum_{k=1}^{K}\Pr(z=k|x_i,\Theta_0)=1
\]
So it eventuates that
\[
\sum_{z_1=1}^{K}\cdots\sum_{z_{N-1}=1}^{K}\prod_{j=1}^{N-1} h(z_j|x_j)=
\sum_{z_1=1}^{K}\cdots\sum_{z_{N-2}=1}^{K}\prod_{j=1}^{N-2}h(z_j|x_j)
\]
and, by reverse induction on $z_n$ for $n=N-2,N-3,\ldots,2,1$ it follows (somewhat miraculously) that
\[
\sum_{z_1=1}^{K}\cdots\sum_{z_{N-1}=1}^{K}\prod_{j=1}^{N-1} h(z_j|x_j)=
\sum_{z_1=1}^{K}h(z_1|x_1)=1
\]
Returning to (\ref{baum3}), we now have
\begin{eqnarray}
Q(\Theta,\Theta_0) &=& \sum_{z_1=1}^{K}g(z_1,x_1)h(z_1|x_1)+\cdots+\sum_{z_N=1}^{K}g(z_N,x_N)h(z_N|x_N)\nonumber\\
 &=& \sum_{n=1}^{N}\sum_{k=1}^{K}g(z=k,x_n)h(z=k|x_n)\nonumber\\
 &=& \sum_{n=1}^{N}\sum_{k=1}^{K}\Pr(z=k|x_n,\Theta_0)\log{\rm p}(z=k,x_n|\Theta)\label{baum4}
\end{eqnarray}
Examining the second term in the double sum, we notice that
\begin{eqnarray*}
\log{\rm p}(z=k,x_n|\Theta) &=& \log\left(\Pr(z=k|\Theta){\rm p}(x_n|z=k,\Theta)\right)\\
 &=& \log \pi_k+\log {\rm p}(x_n|z=k,\Theta)
\end{eqnarray*}
Applying Bayes' rules to the first term in (\ref{baum4}) we can show that
\[
\Pr(z=k|x_n,\Theta_0)=\frac{\pi^{(0)}_k{\rm p}(x_n|\theta^{(0)}_k)}{
\sum_{j=1}^{K}\pi^{(0)}_j{\rm p}(x_n|\theta^{(0)}_j)}
\]
where we have defined $\theta^{(p)}_k=\{\pi^{(p)}_k, \mu^{(p)}_k,P^{(p)}_k\}$ as the set of parameters for mixture component $k$ at pass $p$ and ${\rm p}(x|\theta_j)$ is shorthand for the PDF of component $j$ of the mixture density and the superscript denotes quantities computed at the relevant pass. From (\ref{baum4}), the E-step for the EM algorithm is therefore given for a general finite mixture as
\begin{equation}\label{estep}
Q(\Theta,\Theta_{p-1})=\sum_{n=1}^{N}\sum_{k=1}^{K}w_{nk}(\Theta_{p-1})\left(\log\,\pi_k+\log\,{\rm p}(x_n|\theta_k)
\right)
\end{equation}
where the ``weights'' defined for $n=1,\ldots,N$ and $k=1,\ldots,K$ are given by
\[
w_{nk}(\Theta_{p-1})=
\frac{\pi^{(p-1)}_k{\rm p}(x_n|\theta^{(p-1)}_k)}{
\sum_{j=1}^{K}\pi^{(p-1)}_j{\rm p}(x_n|\theta^{(p-1)}_j)}
\]
In the Gaussian mixture case, the weights are obtained as
\[
w_{nk}(\Theta_{p-1})=\frac{\pi^{(p-1)}_k {\rm N}(x_n;\mu^{(p-1)}_k,P^{(p-1)}_k)}{\sum_{j=1}^{K}\pi^{(p-1)}_j{\rm N}(x_n;\mu^{(p-1)}_j,P^{(p-1)}_j)}
\]
where the PDF was defined in (\ref{gaussian}). In this case, the E-step takes the form
\begin{eqnarray*}
Q(\Theta,\Theta_{p-1}) &=& \sum_{n=1}^{N}\sum_{k=1}^{K}w_{nk}(\Theta_{p-1})\left(\log\,\pi_k-\frac{n_x}{2}\log(2\pi)-\hlf\log({\rm det}P_k)\right.\\
 &&~ \left.-\hlf(x_n-\mu_k)^TP_k^{-1}(x_n-\mu_k)\right)
\end{eqnarray*}

The updated mixture weights, means and covariance matrices are obtained from the M-step, subject to the constraint that the mixture weights sum to unity. The update for the mixture weights is easy to derive and does not depend on the type of mixture PDF. The updates for the mixture means and covariances are more laborious. (Bilmes \cite{Bilmes} has detailed derivations, and similar calculations appear in section \ref{skemdetails}.) The result is:
\begin{eqnarray*}
\pi^{(p)}_k &=& \frac{1}{N}\sum_{n=1}^{N}w_{nk}(\Theta_{p-1}) \\
\mu^{(p)}_k &=& \frac{\sum_{n=1}^{N}w_{nk}(\Theta_{p-1})x_n}{\sum_{n=1}^{N}w_{nk}(\Theta_{p-1})}\\
P^{(p)}_k &=& \frac{\sum_{n=1}^{N}w_{nk}(\Theta_{p-1})(x_n-\mu^{(p)}_k)(x_n-\mu^{(p)}_k)^T}{\sum_{n=1}^{N}w_{nk}(\Theta_{p-1})}
\end{eqnarray*}
The loop of the EM algorithm is completed by setting $\Theta_p=\{\pi^{(p)}_k, \mu^{(p)}_k,P^{(p)}_k\}_{k=1}^K$ and $p\leftarrow p+1$ and iterating until convergence. Note that the number of mixture components $K$ must be chosen beforehand. Numerical examples of a 2-D simulation of the EM algorithm, showing the effect of the number of mixture components, can be found in \cite{Pulford2020b}.

\section{Shared Kernel EM Algorithm}\label{skem}
\subsubsection*{Notation}
The treatment of the SKEM algorithm uses the following notation, which is consistent with the notation in section \ref{EMalg}.
\begin{itemize}
\item Incomplete data $X=\{x_1,\ldots,x_N\},~x_i\in\Reals^M$, also called unlabelled or uncategorised data.
\item Class labels $C=\{c_1,\ldots,c_N\}$, with each element $c_i\in\{1,\ldots,L\}$, where $L$ is the number of classes.\footnote{In the algorithm implementation section we use NC for the number of classes.}
\item Missing data $Z=\{z_1,\ldots,z_N\}$, also called hidden or latent variables. $z_n=k$ indicates that $x_n$ is associated with mixture component $k$.
\item Complete data $Y=(X,Z)$, the union of the incomplete and missing data.
\item Shared kernel parameters for component $k$ in a mixture $\theta_k$.
\item Full set of parameters $\Theta$, including weights.
\item Estimated parameters $\Theta_p$ at pass $p$ of the algorithm.
\end{itemize}
In the case of a $K$-component multivariate Gaussian shared kernel, the full parameter set $\Theta$ comprises the matrix of class-conditioned weights $\{\pi_{kj}\}$, $k=1,\ldots,K$, $j=1,\ldots,L$, mean vectors $\{\mu_k\}$, where $\mu_k\in\Reals^M$, and covariance matrices $\{P_k\}$, where $P_k\in\Reals^{M\times M}$. The shared kernel parameters for mixture component $k$ are $\theta_k=\{\mu_k,P_k\}$.

\subsection{SKEM Algorithm Details}\label{skemdetails}
The class-conditioned density for the SKEM algorithm, for some $x\in\Reals^M$, is assumed to take the form of a Gaussian mixture with parameters $\Theta$
\begin{equation}\label{pxcj}
{\rm p}(x|c=j,\Theta)=\sum_{k=1}^{K}\pi_{kj}\,{\rm N}\{x;\mu_k,P_k\}
\end{equation}
where we can interpret the weight $\pi_{kj}$ as the probability that a data point (feature vector) is associated with component $k$ of the mixture density conditioned on the class $j$. We will also write the association hypothesis in the form $z_n=k$ for some data point $x_n$.

The objective is to determine the maximum likelihood estimate of the shared kernel density parameters and weights such that:
\begin{equation}\label{em2}
    \hat{\Theta}=\arg\max_{\Theta} {\rm p}(X|C,\Theta)
\end{equation}
which differs from the conventional EM criterion both in the assumed form of the conditional densities (the conditioning is on the mixture component in standard EM), and by the presence of the class labels in the conditioning on the right side of (\ref{em2}). The latter makes the SKEM problem a supervised learning problem. As before, the key to solving the optimisation problem is to introduce so-called hidden variables $Z$ that cover all possible association hypotheses between the data and the mixture components, given the class to which the particular feature vector pertains. The association mechanism is analogous to that used in multiple target tracking approaches like the joint probabilistic data association (JPDA) filter \cite{Fortmann2}, where there is a need to associate measurements with target objects. Given an initial parameter estimate $\Theta_0$, Baum's auxiliary function is defined as:
\begin{eqnarray}\label{baumc}
    Q(\Theta,\Theta_0)&=&{\rm E}_Z[\log {\rm p}(X,Z|C,\Theta) | X,\Theta_0]\label{baum2c}\\
    &=&\sum_{z_1=1}^{K}\cdots\sum_{z_N=1}^{K}\log{\rm p}(X,Z|C,\Theta)\Pr(Z|X,\Theta_0)\label{baum2d}
\end{eqnarray}
As in section \ref{EMalg}, the logarithm of the complete data likelihood decomposes via the {\em iid} assumption on the data into a sum of factors in the E-step. This is maximised in the M-step to yield parameter updating equations. The calculations are similar in style to the standard EM derivation, but more complicated due to the presence of class conditioning, which requires more specialised ``accounting.''
We first state the result below and then outline the proof in the 2-class case, giving a worked example to justify why class-conditioned indexing is needed for a full expression of the algorithm. Many of the subtleties of the calculations, including the class-conditioned indexing, are omitted from \cite{Titsias}.

\subsubsection*{Statement of the SKEM algorithm for Gaussian mixtures}
The class vector $C$ partitions the data set into non-overlapping subsets $X_i$, containing the feature vectors for class $i$, with cardinality $l(i)=|X_i|$ defined by
\begin{equation}\label{skemXi}
X_i=\{x_j,~j\in\Gamma_i\},~i=1,\ldots,L
\end{equation}
where $\Gamma_i$ the index set for class $i$, that is, the set of indexes from $\{1,\ldots,N\}$ corresponding to all the class $i$ feature vectors. Note that the sum of the $l(i)$ is the sum of the numbers of feature vectors in each of the $L$ classes, namely $l(1)+\cdots+l(L)=N$. The missing data (latent variables) $Z$ are also partitioned as
\begin{equation}\label{skemZi}
Z_i=\{z_j,~j\in\Gamma_i\},~i=1,\ldots,L
\end{equation}
It is convenient to write the elements of  $\Gamma_i$ in the following way
\begin{equation}\label{skemGamma}
\Gamma_i=\{n_{ij}\},~i=1,\ldots,L,~j=1,\ldots,l(i)
\end{equation}
Clearly, every feature vector belongs to one and only one class, so we have
\[
\bigcup_{i=1}^{L}\Gamma_i=\{1,\ldots,N\},~\Gamma_i\cap\Gamma_j=0,~i\neq j
\]
Let $\Theta_0=\{\pi^{(0)}_{kj}, \mu^{(0)}_k,P^{(0)}_k\}_{k=1\!,\,j=1}^{K,~~L}$, be an initial estimate of the shared kernel model parameters. The SKEM algorithm comprises the following iterative steps, performed in order, starting with $p=1$.
\begin{eqnarray}
w^{(p)}_{n_{ij}k}&=&\frac{\pi^{(p-1)}_{ki} {\rm N}(x_{n_{ij}};\,\mu^{(p-1)}_k,P^{(p-1)}_k)}{\sum_{k=1}^{K}\pi^{(p-1)}_{ki}{\rm N}(x_{n_{ij}};\,\mu^{(p-1)}_k,P^{(p-1)}_k)} \label{skemw}\\
\pi^{(p)}_{ki} &=& \frac{1}{l(i)}\sum_{n\in\Gamma_i}w^{(p)}_{nk} \label{skempi}\\
\mu^{(p)}_k &=& \frac{\sum_{n=1}^{N}w^{(p)}_{nk}x_n}{\sum_{n=1}^{N}w^{(p)}_{nk}} \label{skemmu}\\
P^{(p)}_k &=& \frac{\sum_{n=1}^{N}w^{(p)}_{nk}(x_n-\mu^{(p)}_k)(x_n-\mu^{(p)}_k)^T}{\sum_{n=1}^{N}w^{(p)}_{nk}} \label{skemP}\\
p&\leftarrow& p+1\nonumber
\end{eqnarray}
(The presence of indexes $n_{ij}$ and $n$ in the equations is not a typographical error!)
At all iterations, the class-conditioned mixture weights $\pi_{ki}$ and posterior weights (association probabilities) $w_{nk}$ satisfy the normalisation conditions
\begin{equation}\label{skemnorm}
\sum_{k=1}^{K}\pi_{ki}=1,~ \sum_{k=1}^{K}w_{nk}=1.
\end{equation}

\subsection*{Proof Outline}
We briefly recall the estimation problem in question: given training data $\{X,C\}$ consisting of an $M\times N$ matrix of data and corresponding class labels $C$, obtain maximum likelihood parameters estimates satisfying (\ref{em2}) when a model of the form (\ref{pxcj}) applies to the individual feature vectors $x_n$. The SKEM provides the solution by maxmising Baum's auxiliary function in (\ref{baumc}). The proof is split into two parts. The first part deals with the E-step, the second with the M-step.

\subsubsection*{Evaluation of Baum's Auxiliary Function for the SKEM}
We start by expressing the joint likelihood function for the complete data $Y=(X,Z)$ by partitioning the data set according to the class $c_i$. On account of the iid assumption on the feature vectors, and bearing in mind the definitions in (\ref{skemXi}), (\ref{skemZi}) and (\ref{skemGamma}), we have
\begin{eqnarray}
{\rm p}(X,Z|C,\Theta) &=& \prod_{i=1}^{L}\Pr(Z_i|c_i,\Theta)\,{\rm p}(X_i|Z_i,c_i,\Theta)\nonumber\\
 &=& \prod_{i=1}^{L}\prod_{j=1}^{l(i)}\Pr(z_{n_{ij}}|c_i,\Theta)\,{\rm p}(x_{n_{ij}}|Z_i,c_i,\Theta)\\
 &=& \prod_{i=1}^{L}\prod_{j=1}^{l(i)}\Pr(k_{ij}|c_i,\Theta)\,{\rm p}(x_{n_{ij}}|\theta_k)
\end{eqnarray}
where, for subsequent notational convenience, we have defined $k_{ij}\keiko z_{n_{ij}}$, being the component in the mixture assigned to feature vector $x_{n_{ij}}$. The notation ${\rm p}(x_{n_{ij}}|\theta_k)$ is shorthand for ${\rm N}\{x;\mu_k,P_k\}$ as defined in (\ref{gaussian}). Note that the decomposition into products {\em not} involving sums of terms dependent on the parameters is crucial to the simplification of the log likelihood. So taking the logarithm gives a sum of simple terms:
\begin{equation}\label{logpxzc}
\log {\rm p}(X,Z|C,\Theta)=\sum_{i=1}^{L}\sum_{j=1}^{l(i)}\log\left(\Pr(k_{ij}|c_i,\Theta)\,{\rm p}(x_{n_{ij}}|\theta_k)\right)
\end{equation}
At this point we make a number of definitions that will be useful in the sequel.
\begin{eqnarray}
\pi_{kc} &=& \Pr(z=k|c,\Theta),~k=1:K,~c=1:L\label{pikc}\\
g_{ij}&\keiko& g(k_{ij},x_{n_{ij}})= \log\Pr(k_{ij}|c_i,\Theta)+\log{\rm p}(x_{n_{ij}}|\theta_k)\label{gij}\\
h_{ij}&\keiko& h(k_{ij}|x_{n_{ij}})=\Pr(k_{ij}|x_{n_{ij}},c_i,\Theta_0),~i=1:L,~j=1:l(i)\label{hij}
\end{eqnarray}
Since both $\pi_{kc}$ and $h_{ij}$ are discrete probability distributions (probability mass functions, PMF), they satisfy the normalisation conditions
\begin{equation}\label{norm}
\sum_{k=1}^{K}\pi_{kc}=1,~\sum_{k_{ij}=1}^{K}h(k_{ij}|x_{n_{ij}})=1
\end{equation}
The reason for introducing the $h_{ij}$ functions is to express the joint PMF of the latent variables given the data and initial parameter estimates, {\em viz.}:
\begin{eqnarray}
\Pr(Z|X,\Theta_0) &=& \prod_{i=1}^{L}\Pr(Z_i|X_i,\Theta_0) = \prod_{i=1}^{L}\prod_{j=1}^{l(i)}\Pr(z_{n_{ij}}|x_{n_{ij}},c_i,\Theta_0)\nonumber\\
&=&\prod_{i=1}^{L}\prod_{j=1}^{l(i)}h(k_{ij}|x_{n_{ij}})\keiko\prod_{i=1}^{L}\prod_{j=1}^{l(i)}h_{ij}\label{pzxtheta0}
\end{eqnarray}
The construction of Baum's auxiliary function requires the expectation over the {\em sequence} of latent variables to be calculated. Since the latter are discrete, an $N$-fold sum over the components $z_i$ of $Z$ is involved. Each $z_i$ can take any of $K$ values; thus there are $K^N$ terms in the expectation in equation (\ref{baum2d}). The class conditioning allows this $N$-fold sum to be rearranged as $L$ groups of sums each having $l(i)$ individual sums. This corresponds to the factorisation $K^{l(1)}\times\cdots K^{l(L)}=K^N$. With the preceding notational definitions and equations (\ref{logpxzc}) and (\ref{pzxtheta0}), it is now possible to write $Q(\Theta,\Theta_0)$ as
\begin{equation}
Q(\Theta,\Theta_0)=
\underbrace{\left(\sum_{k_{11}=1}^{K}\cdots\sum_{k_{1l(1)}=1}^{K}\right)}_{l(1)\mbox{ sums}}
\cdots
\underbrace{\left(\sum_{k_{L1}=1}^{K}\cdots\sum_{k_{Ll(L)}=1}^{K}\right)}_{l(L)\mbox{ sums}}
\sum_{i=1}^{L}\sum_{j=1}^{l(i)}g_{ij}
\prod_{m=1}^{L}\prod_{r=1}^{l(m)}h_{mr}\label{Lsums}
\end{equation}
Rather than deal with the general case, we present the simplification of the result for the $L=2$ class case. The general case can be dealt with in an analogous manner but is more tedious, and is therefore omitted. Setting $L=2$ in (\ref{Lsums}), we get
\begin{equation}
Q(\Theta,\Theta_0)=
\underbrace{\left(\sum_{k_{11}=1}^{K}\cdots\sum_{k_{1l(1)}=1}^{K}\right)}_{l(1)\mbox{ sums}}
\underbrace{\left(\sum_{k_{21}=1}^{K}\cdots\sum_{k_{2l(2)}=1}^{K}\right)}_{l(2)\mbox{ sums}}
\sum_{i=1}^{2}\sum_{j=1}^{l(i)}g_{ij}
\prod_{m=1}^{2}\prod_{r=1}^{l(m)}h_{mr}\label{Lsums2}
\end{equation}
Now define
\[
T_{ij}=\underbrace{\left(\sum_{k_{11}=1}^{K}\cdots\sum_{k_{1l(1)}=1}^{K}\right)}_{l(1)\mbox{ sums}}
\underbrace{\left(\sum_{k_{21}=1}^{K}\cdots\sum_{k_{2l(2)}=1}^{K}\right)}_{l(2)\mbox{ sums}}
g_{ij}\prod_{m=1}^{2}\prod_{r=1}^{l(m)}h_{mr}
\]
so that
\[
Q(\Theta,\Theta_0)=\sum_{i=1}^{2}\sum_{j=1}^{l(i)}T_{ij}
\]
Consider the first term on the right hand side, and re-express it as
\[
T_{11}=\sum_{k_{11}}g_{11}h_{11}
\underbrace{\left(\sum_{k_{12}}\cdots\sum_{k_{1l(1)}}\right)}_{l(1)-1\mbox{ sums}}
\underbrace{\left(\sum_{k_{21}}\cdots\sum_{k_{2l(2)}}\right)}_{l(2)\mbox{ sums}}
\prod_{(m,r)\neq(1,1)}h_{mr}
\]
in which for notational convenience we have omitted the limits in the sums. We treat the right hand part of $T_{11}$ (after the term $g_{11}h_{11}$) in a similar manner to the proof of the standard EM algorithm E-step:
\begin{eqnarray*}
 && \underbrace{\left(\sum_{k_{12}}\cdots\sum_{k_{1l(1)}}\right)}_{l(1)-1\mbox{ sums}}
\underbrace{\left(\sum_{k_{21}}\cdots\sum_{k_{2l(2)-1}}\right)}_{l(2)-1\mbox{ sums}}
\left(\sum_{k_{2l(2)}}h_{2l(2)}\right)
\prod_{{(m,r)\neq(1,1)}\atop{(m,r)\neq(2,l(2))}}h_{mr} \\
 &=& \underbrace{\left(\sum_{k_{12}}\cdots\sum_{k_{1l(1)}}\right)}_{l(1)-1\mbox{ sums}}
\underbrace{\left(\sum_{k_{21}}\cdots\sum_{k_{2l(2)-1}}\right)}_{l(2)-1\mbox{ sums}}
\prod_{{(m,r)\neq(1,1)}\atop{(m,r)\neq(2,l(2))}}h_{mr}
\end{eqnarray*}
where we used the normalisation condition (\ref{norm}). Proceeding by induction on both indexes $k_{1j}$ and $k_{2j}$, we obtain
\[
\underbrace{\left(\sum_{k_{12}}\cdots\sum_{k_{1l(1)}}\right)}_{l(1)-1\mbox{ sums}}
\underbrace{\left(\sum_{k_{21}}\cdots\sum_{k_{2l(2)}}\right)}_{l(2)\mbox{ sums}}
\prod_{(m,r)\neq(1,1)}h_{mr}=1
\]
which means that
\[
T_{11}=\sum_{k_{11}=1}^{K}g_{11}h_{11}=\sum_{k_{11}=1}^{K}g(k_{11},x_{n_{11}})\,h(k_{11}|x_{n_{11}})
\]
It follows immediately that
\[
T_{ij}=\sum_{k_{ij}=1}^{K}g_{ij}h_{ij}=\sum_{k_{ij}=1}^{K}g(k_{ij},x_{n_{ij}})\,h(k_{ij}|x_{n_{ij}})
\]
whence
\[
Q(\Theta,\Theta_0)=\sum_{i=1}^{2}\sum_{j=1}^{l(i)}\sum_{k_{ij}=1}^{K}g(k_{ij},x_{n_{ij}})\,h(k_{ij}|x_{n_{ij}})
\]
and, in the general $L$-class case:
\begin{equation}\label{baum2e}
Q(\Theta,\Theta_0)=\sum_{i=1}^{L}\sum_{j=1}^{l(i)}\sum_{k_{ij}=1}^{K}g(k_{ij},x_{n_{ij}})\,h(k_{ij}|x_{n_{ij}})
\end{equation}
In light of definitions (\ref{pikc})--(\ref{hij}), a few tweaks can be made to tidy up this last expression for the auxiliary function. Concentrating on $h_{ij}$, we apply Bayes' rule to obtain
\[
\Pr(k_{ij}|x_{n_{ij}},c_i,\Theta_0)=\frac{\Pr(k_{ij}|c_i,\Theta_0)\,{\rm p}(x_{n_{ij}}|k_{ij},\Theta_0)}
{\sum_{k_{ij}=1}^{K}\Pr(k_{ij}|c_i,\Theta_0)\,{\rm p}(x_{n_{ij}}|k_{ij},\Theta_0)}
\]
where, in the Gaussian mixture case,
\[
{\rm p}(x_{n_{ij}}|k_{ij},\Theta_0)=
{\rm N}(x_{n_{ij}};\,\mu^{0}_{k_{ij}},P^{0}_{k_{ij}})
\]
Upon replacing the index $k_{ij}$ by $k$, reintroducing $\pi_{kc}$ and redefining $h(k|x_{n})$
as $w_{nk}$, we obtain
\begin{equation}\label{wnijk}
w_{n_{ij}k}=\frac{\pi^{0}_{ki}\,{\rm N}(x_{n_{ij}};\,\mu^{0}_k,P^{0}_k)}{\sum_{k=1}^{K}\pi^{0}_{ki}\,{\rm N}(x_{n_{ij}};\,\mu^{0}_k,P^{0}_k)}
\end{equation}
which we recognise as equation (\ref{skemw}). For Baum's auxiliary function in (\ref{baum2d}) we finally obtain the simplified expression
\begin{equation}\label{baums}
Q(\Theta,\Theta_0)=\sum_{i=1}^{L}\sum_{j=1}^{l(i)}\sum_{k=1}^{K} w_{n_{ij}k} % w^{(0)} deleted
\left(\log\pi_{ki}+\log{\rm N}(x_{n_{ij}};\mu_k,P_k)\right)
\end{equation}
This is equivalent to equation (15) in \cite{Titsias}, although the derivation of the auxiliary function does not appear in that work since the explicit calculation of the expectation over {\em all sequences} of latent variables $Z$ is absent. It can be appreciated from the preceding arguments that passing from (\ref{baum2d}) to (\ref{baums}) is not a trivial matter. Furthermore, in \cite{Titsias}, there is an indexing error on the feature vectors, which run repeatedly from $1:N_k$ (or $1:l(i)$ in our notation). However, the feature vector index $n$ should depend on both the class $i$ and the index $j$ within the class-conditioned set of $l(i)$ elements. This reasoning will be made clearer from the worked example in section \ref{wex}.

\subsubsection*{Optimisation of Baum's Auxiliary Function for the SKEM}
In this section we compute the M-step for the SKEM algorithm, having already expressed Baum's auxiliary function $Q(\Theta,\Theta_0)$ in (\ref{baums}). We retain the simpler notation regarding the pass or iteration, with $\Theta$ denoting parameters at the current pass and $\Theta_0$ for parameters from the previous pass. The optimisation problem for the M-step is:
\begin{eqnarray}
 &&\max Q(\Theta,\Theta_0) \label{Mstep}\\
 && \mbox{subject to } \sum_{k=1}^{K}\pi_{kc}=1\nonumber
\end{eqnarray}
The equality-constrained optimisation problem is first expressed as an unconstrained problem by introducing Lagrange multipliers $\lambda=(\lambda_1,\ldots,\lambda_L)$, one per constraint on the parameters to be estimated:
\[
\max_{\Theta,\lambda_1,\ldots,\lambda_L} \calL(\Theta,\lambda_1,\ldots,\lambda_L;\Theta_0)=
Q(\Theta,\Theta_0)-\sum_{i=1}^{L}\lambda_i\left(\sum_{k=1}^{K}\pi_{ki}-1\right)
\]
First-order necessary conditions for a maximum require $\nabla_{\Theta}\calL={\bf 0}$ and $\nabla_{\lambda}\calL={\bf 0}$. Considering first the class-conditioned weights (which do not depend on the form of the conditional PDFs), we must have
\[
\frac{\partial \calL}{\partial\pi_{ki}}=\frac{1}{\pi_{ki}}\sum_{j=1}^{l(i)}w_{n_{ij}k}-\lambda_i=0
\]
and also
\[
\frac{\partial \calL}{\partial\lambda_i}=0~\implies\sum_{j=1}^{l(i)}\sum_{k=1}^{K} w_{n_{ij}k}=\lambda_i
\]
which, because of the normalisation condition in (\ref{skemnorm}), gives $\lambda_i=l(i)$. It follows that
\[
\pi_{ki}=\frac{1}{l(i)}\sum_{j=1}^{l(i)}w_{n_{ij}k}
\]
Recalling the definition of the index set $\Gamma_i$ from (\ref{skemGamma}), this is more naturally expressed as
\begin{equation}
\pi_{ki}=\frac{1}{l(i)}\sum_{n\in\Gamma_i}w_{nk}
\end{equation}
which we recognise as the update equation for $\pi_{ki}$ in (\ref{skempi}).

The update equations for $\mu_k$ and $P_k$, being vector and matrix based, require more effort. We note that the objective function in (\ref{Mstep}) is scalar, so the full machinery of matrix differential calculus \cite{Magnus} is not required. However, we do need the definition of the derivative of a scalar function $\phi(X)$ with respect to a matrix argument $X=(x_{ij})$:
\[
\frac{\partial\phi(X)}{\partial X}=\left(\frac{\partial\phi(X)}{\partial x_{ij}}\right)
\]
We also borrow some results from \cite{Petersen} (section 8.4) concerning the derivatives of the logarithm of a multivariate Gaussian PDF with respect to its mean vector and covariance matrix:
\begin{eqnarray*}
\frac{\partial}{\partial\mu}\log{\rm N}\{x;\,\mu,P\} &=& P^{-1}(x-\mu)\\
\frac{\partial}{\partial P}\log{\rm N}\{x;\,\mu,P\} &=& \hlf P^{-1}
(x-\mu)(x-\mu)^TP^{-1}-\hlf P^{-1}
\end{eqnarray*}
Note that the last equation includes the derivative of the normalisation term of the Gaussian PDF, given below for completeness:
\[
\frac{\partial}{\partial P}\log|\det P|=(P^T)^{-1}=P^{-1}
\]
With these results in hand, the remaining update equations are easy to obtain. For the mean vectors:
\[
\frac{\partial \calL}{\partial\mu_k}=\sum_{i=1}^{L}\sum_{j=1}^{l(i)} w_{n_{ij}k}P^{-1}_k(x_{n_{ij}}-\mu_k)
={\bf 0}
\]
Now pre-multiply by $P_k$ and solve for $\mu_k$
\[
\mu_k = \frac{\sum_{i=1}^{L}\sum_{j=1}^{l(i)} w_{n_{ij}k}\,x_{n_{ij}} }{\sum_{i=1}^{L}\sum_{j=1}^{l(i)} w_{n_{ij}k}}
\]
Noting that $\sum_{i=1}^{L}\sum_{j=1}^{l(i)}f(n_{ij})=\sum_{n=1}^{N}f(n)$, the latter is the same as (\ref{skemmu}).

For the covariance matrices:
\[
\frac{\partial \calL}{\partial P_k}={\bf 0}~\implies\sum_{i=1}^{L}\sum_{j=1}^{l(i)} w_{n_{ij}k}
P^{-1}_k\left[(x_{n_{ij}}-\mu_k)(x_{n_{ij}}-\mu_k)^T P^{-1}_k-I\right]={\bf 0}
\]
Multiply the above by $P_k$ on the left and right to give
\[
\sum_{i=1}^{L}\sum_{j=1}^{l(i)} w_{n_{ij}k}
\left[(x_{n_{ij}}-\mu_k)(x_{n_{ij}}-\mu_k)^T-P_k\right]={\bf 0}
\]
which yields
\[
P_k = \frac{\sum_{i=1}^{L}\sum_{j=1}^{l(i)} w_{n_{ij}k}\,(x_{n_{ij}}-\mu_k)(x_{n_{ij}}-\mu_k)^T }{\sum_{i=1}^{L}\sum_{j=1}^{l(i)} w_{n_{ij}k}}
\]
Rejigging the indexing in the previous equation gives the covariance update in (\ref{skemP}).

\subsection{Worked Example}\label{wex}
A low order example, presented next, serves to clarify the derivation of the SKEM algorithm. Suppose there are $L=2$ classes, $N=5$ feature vectors and $K=2$ components in the mixture. Suppose further that the class label vector is $C=[1, 1, 2, 1, 2]$.
The class vector defines two subsets of feature vectors: $X_1=(x_1,x_2,x_4)$ and $X_2=(x_3,x_5)$, as well as their cardinalities $N_1=3$ and $N_2=2$ satisfying $N_1+N_2=5$.

Using the simplified expression for Baum's auxiliary function (\ref{baums}), and writing $p(x;\theta_k)$ as shorthand for ${\rm N}\{x;\,\mu_k,P_k\}$, we have

\begin{eqnarray*}
Q(\Theta,\Theta_0)
&=&w_{11}(\ln\pi_{11}+\ln p(x_1;\theta_1))+w_{12}(\ln\pi_{21}+\ln p(x_1;\theta_2))\\
&&+w_{21}(\ln\pi_{11}+\ln p(x_2;\theta_1))+w_{22}(\ln\pi_{21}+\ln p(x_2;\theta_2))\\
&&+w_{41}(\ln\pi_{11}+\ln p(x_4;\theta_1))+w_{42}(\ln\pi_{21}+\ln p(x_4;\theta_2))\\
&&+w_{31}(\ln\pi_{12}+\ln p(x_3;\theta_1))+w_{32}(\ln\pi_{22}+\ln p(x_3;\theta_2))\\
&&+w_{51}(\ln\pi_{12}+\ln p(x_5;\theta_1))+w_{52}(\ln\pi_{22}+\ln p(x_5;\theta_2))
\end{eqnarray*}

The SKEM seeks to obtain the shared kernel parameters and class-conditioned weights by optimisation of Baum's auxiliary function with respect to $\Theta$, subject to the normalisation constraints on the weights, {\em viz.}:
\begin{eqnarray*}
 &&\max Q(\Theta,\Theta_0) \\
 && \mbox{subject to } \pi_{11}+\pi_{21}=1\mbox{ and }\pi_{12}+\pi_{22}=1
\end{eqnarray*}
As in the previous section, we introduce Lagrange multipliers, one per constraint:
\[
\max_{\Theta,\lambda_1,\lambda_2} \calL(\Theta,\lambda_1,\lambda_2;\Theta_0)=Q(\Theta,\Theta_0)-\lambda_1(\pi_{11}+\pi_{21}-1)-\lambda_2(\pi_{12}+\pi_{22}-1)
\]
For a maximum, we require $\nabla_{\Theta}\calL={\bf 0}$, $\partial \calL/\partial{\lambda_1}=0$, and $\partial \calL/\partial{\lambda_2}=0$. Considering first the class-conditioned weights and Lagrange multipliers, we have
\begin{eqnarray*}
0 &=& \frac{\partial \calL}{\partial\pi_{11}}=\frac{w_{11}+w_{21}+w_{41}}{\pi_{11}}-\lambda_1 \\
0 &=& \frac{\partial \calL}{\partial\pi_{21}}=\frac{w_{12}+w_{22}+w_{42}}{\pi_{21}}-\lambda_1 \\
0 &=& \frac{\partial \calL}{\partial\pi_{12}}=\frac{w_{31}+w_{51}}{\pi_{12}}-\lambda_2 \\
0 &=& \frac{\partial \calL}{\partial\pi_{22}}=\frac{w_{32}+w_{52}}{\pi_{22}}-\lambda_2 \\
0 &=& \frac{\partial \calL}{\partial\lambda_1}~\implies~ \lambda_1=w_{11}+w_{21}+w_{41}+w_{12}+w_{22}+w_{42}=3\\
0 &=& \frac{\partial \calL}{\partial\lambda_2}~\implies~ \lambda_2=w_{31}+w_{51}+w_{32}+w_{52}=2
\end{eqnarray*}
in which the simplification on the right hand side results from the normalisation constraint on the $w_{nk}$ in (\ref{skemnorm}). We therefore obtain
\begin{eqnarray*}
\pi_{11} &=& {\scriptstyle \frac{1}{3}}(w_{11}+w_{21}+w_{41})\\
\pi_{21} &=& {\scriptstyle \frac{1}{3}}(w_{12}+w_{22}+w_{42})\\
\pi_{12} &=& {\scriptstyle \frac{1}{2}}(w_{31}+w_{51})\\
\pi_{22} &=& {\scriptstyle \frac{1}{2}}(w_{32}+w_{52})
\end{eqnarray*}

The updates for the other parameters clearly depend on the form of the PDFs $p(x;\theta_k)$. In the scalar Gaussian case, denoting the square of the standard deviation as $s=\sigma^2$, the required derivatives are given by
\begin{eqnarray*}
\frac{\partial\ln p(x_j;\theta_i)}{\partial\mu_i} &=& \frac{x_j-\mu_i}{s_i}\\
\frac{\partial\ln p(x_j;\theta_i)}{\partial s_i} &=& \frac{1}{2s_i}\left[\frac{(x_j-\mu_i)^2}{s_i}-1\right]
\end{eqnarray*}
leading to
\begin{eqnarray*}
0\!\!\!\!&=&\!\!\!\! \frac{\partial \calL}{\partial\mu_1}=w_{11}\frac{x_1-\mu_1}{s_1}+w_{21}\frac{x_2-\mu_1}{s_1}+w_{41}\frac{x_4-\mu_1}{s_1}+w_{31}\frac{x_3-\mu_1}{s_1}+w_{11}\frac{x_5-\mu_1}{s_1}\\
0\!\!\!\!&=&\!\!\!\! \frac{\partial \calL}{\partial\mu_2}=w_{12}\frac{x_1-\mu_2}{s_2}+w_{22}\frac{x_2-\mu_2}{s_2}+w_{42}\frac{x_4-\mu_2}{s_2}+w_{32}\frac{x_3-\mu_2}{s_2}+w_{52}\frac{x_5-\mu_2}{s_2}
\end{eqnarray*}
The update equations for the mixture means are therefore
\begin{eqnarray*}
\mu_1 &=& \frac{w_{11}x_1+w_{21}x_2+w_{41}x_4+w_{31}x_3+w_{51}x_5}{w_{11}+w_{21}+w_{41}+w_{31}+w_{51}}\\
\mu_2 &=& \frac{w_{12}x_1+w_{22}x_2+w_{42}x_4+w_{32}x_3+w_{52}x_5}{w_{12}+w_{22}+w_{42}+w_{32}+w_{52}}
\end{eqnarray*}
Focussing on the ``variance'' terms we have
\begin{eqnarray*}
0=\frac{\partial \calL}{\partial s_1}\!\!\!\!&=&\!\!\!\!
\frac{w_{11}}{2s_1}\left[\frac{(x_1-\mu_1)^2}{s_1}-1\right]+
\frac{w_{21}}{2s_1}\left[\frac{(x_2-\mu_1)^2}{s_1}-1\right]+
\frac{w_{41}}{2s_1}\left[\frac{(x_4-\mu_1)^2}{s_1}-1\right]\\
&&+\frac{w_{31}}{2s_1}\left[\frac{(x_3-\mu_1)^2}{s_1}-1\right]+
\frac{w_{51}}{2s_1}\left[\frac{(x_5-\mu_1)^2}{s_1}-1\right]\\
0=\frac{\partial \calL}{\partial s_2}\!\!\!\!&=&\!\!\!\!
\frac{w_{12}}{2s_2}\left[\frac{(x_1-\mu_2)^2}{s_2}-1\right]+
\frac{w_{22}}{2s_2}\left[\frac{(x_2-\mu_2)^2}{s_2}-1\right]+
\frac{w_{42}}{2s_2}\left[\frac{(x_4-\mu_2)^2}{s_2}-1\right]\\
&&+\frac{w_{32}}{2s_2}\left[\frac{(x_3-\mu_2)^2}{s_2}-1\right]+
\frac{w_{52}}{2s_2}\left[\frac{(x_5-\mu_2)^2}{s_2}-1\right]
\end{eqnarray*}
leading to the update equations for the variances
\begin{eqnarray*}
\sigma_1^2\!\!\!\!&=&\!\!\!\!\!
\frac{w_{11}(x_1-\mu_1)^2+w_{21}(x_2-\mu_1)^2+w_{41}(x_4-\mu_1)^2+w_{31}(x_3-\mu_1)^2+w_{51}(x_5-\mu_1)^2}{w_{11}+w_{21}+w_{41}+w_{31}+w_{51}}\\
\sigma_2^2\!\!\!\!&=&\!\!\!\!\!
\frac{w_{12}(x_1-\mu_2)^2+w_{22}(x_2-\mu_2)^2+w_{42}(x_4-\mu_2)^2+w_{32}(x_3-\mu_2)^2+w_{52}(x_5-\mu_2)^2}{w_{12}+w_{22}+w_{42}+w_{32}+w_{52}}
\end{eqnarray*}

\subsection{Shared Kernel Classifier}\label{skemc}
A classifier can be constructed from the family of class-conditioned shared kernel PDFs learned from the data by the SKEM algorithm. The type of classifier depends on the optimisation criterion that one wishes to apply. The most common are the maximum {\em a posteriori} (MAP) and maximum likelihood (ML) criteria. The incomplete data likelihood is:
\[
{\rm p}(x)=\sum_{j=1}^{L}\Pr(c=j)\,{\rm p}(x|c=j)
\]
where there are $L$ classes, $c$ is the class, and the prior class probabilities are $\Pr(c=j)$, $j=1,\ldots,L$. The class-conditioned PDFs for the shared kernel model, in the case of Gaussian kernels, are given by
\begin{equation}\label{pxcj2}
{\rm p}(x|c=j)=\sum_{k=1}^{K}\pi_{kj}\,{\rm N}\{x;\mu_k,P_k\},~\sum_{k=1}^{K}\pi_{kj}=1
\end{equation}
The MAP classifier rule is
\begin{equation}
\hat{c}_{\rm MAP}=\arg\max_{c\in\{1,\ldots,L\}}\Pr(c|x)
\end{equation}
A straightforward application of Bayes' rule yields
\begin{equation}
\hat{c}_{\rm MAP}=\arg\max_{c\in\{1,\ldots,L\}}\frac{\Pr(c)\,{\rm p}(x|c)}{\sum_{j=1}^{L}\Pr(c=j)\,{\rm p}(x|c=j)}
\end{equation}
wherein we can replace the class-conditioned densities by their shared kernel representation in (\ref{pxcj2}).

During training, we can obtain the prior class probabilities as $\Pr(c=j)=l(j)/L$, $j=1,\ldots,L$ where $l(j)=|\Gamma(j)|$ and $\Gamma(j)$ is defined in (\ref{skemXi}). However, in actual practice these prior probabilities are not likely to be representative of a real data set. In the present context, we assume no knowledge of the prior digit probabilities. This is equivalent to taking a uniform prior for the latter. Hence we assume $\Pr(c)=1/L$ for all classes. The MAP criterion therefore reduces to
\begin{equation}
\hat{c}_{\rm MAP}=\arg\max_{c\in\{1,\ldots,L\}}\frac{{\rm p}(x|c)}{\sum_{j=1}^{L}{\rm p}(x|c=j)}
\end{equation}
which is none other than the ML criterion
\begin{equation}
\hat{c}_{\rm ML}=\arg\max_{c\in\{1,\ldots,L\}}{\rm p}(x|c=j)
\end{equation}
In the case of shared kernel PDFs this gives the
\begin{equation}\label{skcl}
\hat{c}_{\rm ML}=\arg\max_{j\in\{1,\ldots,L\}}\sum_{k=1}^{K}\pi_{kj}\,{\rm N}\{x;\mu_k,P_k\}
\end{equation}
which we refer to as the shared kernel classifier (SKC). The SKC is used in section \ref{NS} for the digit recognition problem.

\section{Numerical Simulations}\label{NS}
\subsection{Training and Testing Data}\label{testdata}
In theory, the shared kernel EM algorithm can be applied to data of arbitrary dimension (number of features) subject only to computational constraints. In practice, the SKEM method can only be applied ``as is'' to quite low-dimensional data sets, although we later show how to partially relieve this constraint by deploying the SKEM independently on subsets of features. Rather than choosing a data set whose underlying class-conditioned models are in fact mixture densities of known form, we have opted for a more challenging example based on image recognition of the digits 0 - 9.

Although LeCun's MNIST database \cite{LeCun} of handwritten digit images could be used, it was decided that a synthetic image dataset would provide more flexibility since it can be tailored with a small number of adjustable parameters and be expanded to any size. Like MNIST, the data are split into non-overlapping training and test sets. The training set consists of 100  greyscale 28 x 28 pixel images of each of the 10 digits, which is considerably smaller than the 60,000 in LeCun's data set. The test set contains 250 greyscale 28 x 28 pixel images of each digit.

\begin{figure}
\center
   \includegraphics[height=10cm]{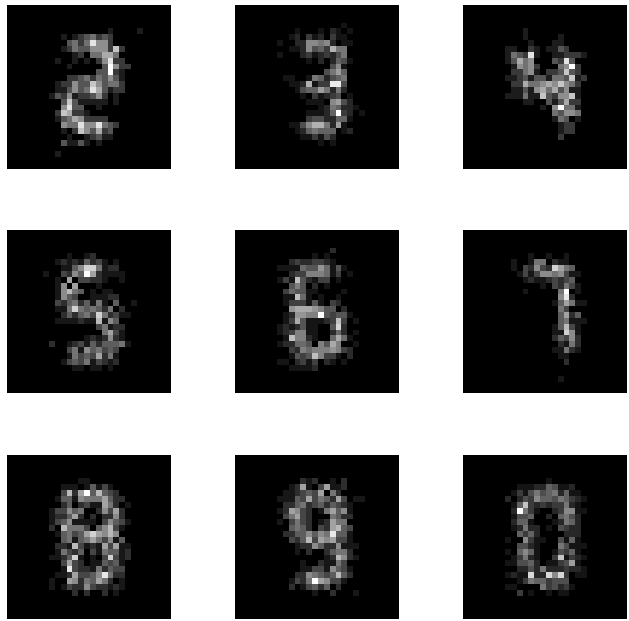}
\caption{Examples of digits 2 - 9 generated using the 7-segment representation.}
\label{figdig}
\end{figure}

The digit representation is based on the so-called ``7-segment'' LED devices used for digital electronic displays in counters and electronic time pieces. The shape of the display is controlled by two parameters $D$ and $d$, which are, respectively, the half-length and half-width of a single segment. The Matlab source code for generating the digits is given in section \ref{7seg}. Examples of the generated digits appear in Fig. \ref{figdig} for $D=2$, $d=0.5$. Further details of the actual data generation process are relegated to the appendix.

We chose a smaller number of training images to demonstrate the efficiency of the SKEM algorithm, that is, its ability to learn an accurate statistical model from a relatively small number of images. The total number of independent parameters in a shared kernel model (with Gaussian mixtures) having $K$ components, $L$ classes and $M$-dimensional features is $MK+KM(M+1)/2+KL$. Thus, assuming, say 10 features, with $K=10$ and $L=10$, there are 750 parameters to be learned from 100 (images) x 10 (digits) x 10 (features) = 10,000 scalar quantities from the training data: a ratio of about 13:1. The larger number of test images enhances the performance assessment by including some of the less frequent types of errors in the confusion matrix, examples of which are given in section \ref{results}.

\subsection{Feature Selection \& Extraction}\label{features}

The shared kernel EM algorithm, like other conventional machine learning algorithms such as support vector machines (SVM) and random forests, requires features to be extracted from the set of training images. This is mainly for computational reasons: even with a small 28 x 28 image, there are 784 individual pixels, which is too large a dimension for conventional machine learning techniques. Note that feature extraction from the greyscale image is not the only way to apply EM algorithms to the digit recognition problem. One can also binarise the image and apply an unsupervised EM algorithm using a mixture of Bernoulli distributions. This  type of approach is illustrated in section 9.3.3 of \cite{Bishop2009} for subset of digits (2, 3 and 4).

Image processing has evolved from basic edge, corner and blob detection along with data reduction steps like principal components analysis (PCA) and canonical variates analysis (CVA), to much more sophisticated approaches like deep learning implemented on massively parallel processor architectures such as GPUs. Deep convolutional neural networks are constrained to self-organise, defining their own feature maps and connectivity on presentation of training images, so there is no need for conventional data reduction or feature extraction. Deep convnets achieve the best performance in terms of classification error rate in the MNIST league table \cite{LeCun2}, but only by a fraction of a percent over conventional feature-based approaches like SVM. K-nearest neighbour classifiers teamed with  non-linear deformation can also achieve sub 1\% error rates on MNIST with advanced feature selection methods \cite{Keysers}.

In order to apply the SKEM algorithm to the digit recognition (or classification) problem, we take as features a principal components analysis (PCA) decomposition of a 28 x 28 greyscale image. The image is first normalised by subtraction of the mean and division by the standard deviation, prior to computation of the PCA using singular value decomposition (SVD).

\begin{figure}
\center
   \includegraphics[width=13cm]{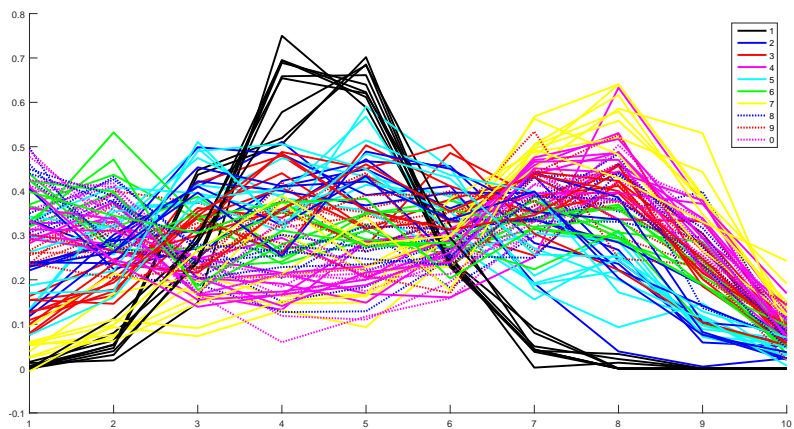}
\caption{First order PCA features for 10 samples of the digits 0 - 9.}
\label{figpca}
\end{figure}

In order to reduce the dimensionality of the feature data, we take only the first column of the PCA decomposition of 28 elements, which corresponds to the transformation vector that yields the maximum data variance among all possible rows. For the 7-segment synthetic images, which are centred on defined segment locations and surrounded by a uniform black background, we can reliably exclude the first 10 and last 8 components, leaving 10 generally non-zero components at indexes 11 to 20. To give some idea of the variability of the PCA features, refer to Fig. \ref{figpca}, which overlays the PCA features for 10 samples of the digits 0 - 9.

After examination of the feature scatter of these 10 components for each digit in a data set with 100 or more samples of each digit, it becomes apparent that there is considerable overlap in feature space for certain digits, such as 6, 8, 9 and 0. To provide better discrimination, a further 4 features are added that correspond the relative ``mass'' in each of 4 quadrants centred at the origin.\footnote{Total greyscale intensity would be a more accurate word than ``mass'' in this instance.} Three of these four quantities are independent, so the total number of independent extracted features is 10 (for PCA features) plus 3 (for quadrant mass), totalling 13. This limits the dimension of the data matrix to $N$ x 13 where $N$ is the total number of training samples. The simulations for the SKEM algorithm considered various subsets of these features, including all 13. The feature extraction code is implemented in the Matlab function calc\_pca\_features() given in section \ref{fextract}.

\subsection{Algorithm Implementation}\label{algsimpl}
The SKEM algorithm described in section \ref{skem} has been implemented in Matlab as a generic algorithm module with a minimal set of input/output parameters. The source code, which is quite compact, is contained in section \ref{srcEM}. Convergence is assessed by testing each class-conditioned log likelihood. Convergence is declared when all of the latter are nearly stationary according to a pre-specified tolerance (taken as 0.1), or a maximum number of passes (NP) has been reached.

As well as the data set X and class labels C, the number of mixture components K is implicit in the initial shared kernel parameters \{Mu, P, Pi\}. The mixture parameters after training are the output of the program. The input specification is below. We use NC instead of $L$ for the number of classes.
\begin{itemize}
\item X -  M-D Data matrix (N x M), N=number of samples, M = number of features (dimension).
\item C - Vector of class labels (same length as X).
\item Mu - Initial estimates of mixture means (M x K)
\item P - Initial estimates of mixture covariance matrices (M x KM) (variance if M = 1).
\item Pi - Initial estimates of mixing probabilities (K x NC).
\item NP - Maximum number of passes.
\end{itemize}
Apart from the choice of the number of mixture components K, which is arguably the most important freely chosen parameter of the algorithm (see section \ref{Kvar}), the initial mixture parameters estimates also have a large bearing on algorithm convergence. The method for choosing the initial values of  \{Mu, P, Pi\} is clearly dimension- and data-dependent. The features for the digit data are normalised PCA coefficients and relative quadrant mass, all of which are between 0 and 1. In the experiments reported in section \ref{results}, the following rules were adopted. For the simulations presented later, the only source of random variation in the Monte Carlo runs is due to different initial mixture mean values.
\begin{enumerate}
\item Initial mixture means Mu are iid uniform [0,1] random variates (in all dimensions and for all components).
\item Initial mixture covariance matrices $P_k$ are taken to be identical and equal to 0.3 $I_M$, $k=1,\ldots,K$, where $I_M$ is an $M\times M$ identity matrix.
\item Initial mixing probabilities are uniform for each of the NC classes.
\end{enumerate}

Additionally, given a data set with N samples and M dimensions (features), one can implement the shared kernel classifier section \ref{skemc} in one of two ways. The brute force way is to take the full dimension M of the feature vector, which can lead to numerical problems due to underflow and overflow. To understand this, it is enough to consider the behaviour of a multi-dimensional Gaussian PDF, which becomes more ``peaked'' as the dimensionality increases due to the PDF normalisation requirement. For instance, in a 9-dimensional version of the SKEM, the maximum multivariate Gaussian PDF evaluation regularly attained $10^{15}$.

Alternatively, the features can be divided into non-overlapping subsets corresponding to a decomposition of the joint data likelihood into ``independent'' factors. While in most cases there is little theoretical justification for this since features are typically dependent (although they may be independent given the class), there are practical benefits (as noted by Luttrell \cite{Luttrell}). For instance, one version of the algorithm calls for a 6-D feature vector $X$ comprising 3 quadrant mass features and 3 PCA features. This can be implemented using two SKEM algorithms both running on 3-D feature vectors ($X_1$ and $X_2$). Suppose the estimated shared kernel parameters are, respectively, \{Mu$_1$, P$_1$, Pi$_1$\} and \{Mu$_2$, P$_2$, Pi$_2$\}. Denote the likelihood for each class $c$, data vector $X_i$ and mixture parameters $i$ as $p_{c}({\rm X}_i;\,{\rm Mu}_i, {\rm P}_i, {\rm Pi}_i)$. The overall likelihood for class $c$ due to all 6 features is then approximated by $p_{c}(X_1;\,{\rm Mu}_1, {\rm P}_1, {\rm Pi}_1)\,p_{c}(X_2;\,{\rm Mu}_2, {\rm P}_2, {\rm Pi}_2)$. The shared kernel classifier in section \ref{skemc} can be applied by substituting the product likelihood in place of the full joint data likelihood in equation (\ref{skcl}). In what follows, we refer to this implementation of the shared kernel classifier as $M_1$-D x $M_2$-D, where $M_1$ and $M_2$ are the dimensions of the features subsets. Obvious generalisations apply to this definition.

With the previous remarks in mind, we can now present the full list of 10 versions of the SKEM algorithm implemented in this report for the digit recognition problem. Results are presented in the following section.

\begin{enumerate}
\item 3-D quadrant mass components 2:4 only (no PCA features);
\item 3x(1-D) quadrant mass components 2:4 only (product of 3 1-D PDFs, with no PCA features);
\item (3+1)=4-D consisting of quadrant mass components 2:4 with PCA component 4;
\item (3+2)=5-D consisting of quadrant mass components 2:4 with PCA components 4 and 5;
\item (2-D)x(3-D) quadrant mass with PCA components 4 and 5 (product of 2-D and 3-D PDFs);
\item (3+3)=6-D consisting of quadrant mass with PCA components 2, 4 and 5;
\item (3-D)x(3-D) quadrant mass with PCA components 2, 4 and 5 (product of 3-D and 3-D PDFs);
\item (6+3)=9-D quadrant mass with PCA components 2:7;
\item 10-D PCA components 1:10 only (not quadrant mass), with zero-mean, $\sigma=0.01$ Gaussian dither on zero feature values;
\item (10+3)=13-D consisting of all 10 PCA components (with dither) and quadrant mass components 2:4.
\end{enumerate}

\subsection{Simulation Results}\label{results}

\begin{table}
\center
\begin{tabular}{|c|c|c|c|c|c|c|c|c|c|c|}
\hline
Algorithm & 3D & 3x1D & 4D & 5D & 2Dx3D & 6D & 3Dx3D & 9D & 10D & 13D\\
\hline
\hline
Med Acc & 0.847 & 0.838 & 0.896 & 0.856 & 0.958 & 0.855 & 0.958 & 0.825 & 0.704 & 0.799\\
\hline
Mean Acc & 0.801 & 0.836 & 0.778 & 0.742 & 0.856 & 0.740 & 0.858 & 0.767 & 0.640 & 0.722\\
\hline
Std Acc & 0.141 & 0.033 & 0.238 & 0.269 & 0.273 & 0.268 & 0.270 & 0.233 & 0.157 & 0.239\\
\hline
Med MIoU & 0.780 & 0.779 & 0.842 & 0.777 & 0.931 & 0.775 & 0.930 & 0.739 & 0.554 & 0.700\\
\hline
Mean MIoU & 0.727 & 0.777 & 0.707 & 0.671 & 0.819 & 0.668 & 0.820 & 0.693 & 0.496 & 0.636\\
\hline
Std MIoU & 0.149 & 0.035 & 0.251 & 0.283 & 0.293 & 0.283 & 0.290 & 0.246 & 0.146 & 0.252\\
\hline
\end{tabular}
\caption{Performance metrics for all algorithms (all 500 runs).}
\label{Tab1}
\end{table}

\begin{table}
\center
\begin{tabular}{|c|c|c|c|c|c|c|c|c|c|c|}
\hline
Med Acc & 2Dx3D & 3Dx3D & 4D & 5D & 6D & 3D & 3x1D & 9D & 13D & 10D\\
\hline
Mean Acc & 3Dx3D & 2Dx3D & 3x1D & 3D & 4D & 9D & 5D & 6D & 13D & 10D\\
\hline
Med MIoU & 2Dx3D & 3Dx3D & 4D & 3D & 3x1D & 5D & 6D & 9D & 13D & 10D\\
\hline
Mean MIoU & 3Dx3D & 2Dx3D & 3x1D & 3D & 4D & 9D & 5D & 6D & 13D & 10D\\
\hline
\end{tabular}
\caption{Algorithm ranking best to worst according to various metrics (all 500 runs).}
\label{Tab2}
\end{table}

\begin{table}
\center
\begin{tabular}{|c|c|c|c|c|c|c|c|c|c|c|}
\hline
Algorithm & 3D & 3x1D & 4D & 5D & 2Dx3D & 6D & 3Dx3D & 9D & 10D & 13D\\
\hline
Med Acc & 0.848 & 0.838 & 0.896 & 0.900 & 0.959 & 0.899 & 0.958 & 0.899 & 0.707 & 0.803\\
\hline
Mean Acc & 0.828 & 0.838 & 0.858 & 0.860 & 0.956 & 0.854 & 0.954 & 0.849 & 0.689 & 0.819\\
\hline
Std Acc & 0.036 & 0.004 & 0.066 & 0.068 & 0.008 & 0.075 & 0.024 & 0.066 & 0.047 & 0.079\\
\hline
Med MIoU & 0.780 & 0.779 & 0.842 & 0.849 & 0.931 & 0.849 & 0.931 & 0.848 & 0.558 & 0.722\\
\hline
Mean MIoU & 0.755 & 0.778 & 0.790 & 0.794 & 0.926 & 0.787 & 0.923 & 0.779 & 0.540 & 0.736\\
\hline
Std MIoU & 0.051 & 0.004 & 0.087 & 0.089 & 0.014 & 0.098 & 0.031 & 0.089 & 0.057 & 0.107\\
\hline
OK runs &   481 &   499 &   445 &   412 &   440 &   416 &   443 &   438 &   450 &   418\\
\hline
\end{tabular}
\caption{Performance metrics for all algorithms with accuracy $>0.5$. Last line gives number of runs accepted.}
\label{Tab3}
\end{table}

\begin{table}
\center
\begin{tabular}{|c|c|c|c|c|c|c|c|c|c|c|}
\hline
Med Acc & 2Dx3D & 3Dx3D & 5D & 6D & 9D & 4D & 3D & 3x1D & 13D & 10D\\
\hline
Mean Acc & 2Dx3D & 3Dx3D & 5D & 4D & 6D & 9D & 3x1D & 3D & 13D & 10D\\
\hline
Med MIoU & 2Dx3D & 3Dx3D & 5D & 6D & 9D & 4D & 3D & 3x1D & 13D & 10D\\
\hline
Mean MIoU & 2Dx3D & 3Dx3D & 5D & 4D & 6D & 9D & 3x1D & 3D & 13D & 10D\\
\hline
\end{tabular}
\caption{Algorithm ranking best to worst according to various metrics for runs with accuracy $>0.5$.}
\label{Tab4}
\end{table}

All classifier performance metrics (e.g. sensitivity and specificity) can be computed from the confusion matrix. We have limited discussion to the mean accuracy and mean intersection over union (IoU) or Jaccard coefficient. For each of the 10 implemented algorithms in section \ref{algsimpl}, the confusion matrix $C^*_{alg}$ obtained for the run with the highest mean accuracy is also provided in section \ref{confmat}. Testing was carried out on 250 samples of each digit, so an ideal 10-class classifier would furnish a 10 x 10 diagonal matrix with 250 for each diagonal element. In each case, the ordering of classes is [1:9,0], with 0 corresponding to 10th element. The source code for the computation is provided in section \ref{confus}.

It is clear from the confusion matrix results that certain implementations of the SKEM algorithm provide no discrimination of certain digits using the supplied features. For instance, the 3-D SKEM, based only on quadrant mass, consistently confuses digit 1 with digit 9, while the 6-D SKEM, which includes PCA components 2, 4 and 5,  confuses digits 1 and 9 with digit 7.

The results for mean accuracy and mean IoU are given in Tables \ref{Tab1} - \ref{Tab4}. These are grouped into two sets of two tables each. Tables \ref{Tab1} and \ref{Tab2} give performance metrics and rankings for the 10 algorithms on the basis of all 500 test runs with no exclusions for ``bad runs,'' in which the SKEM algorithm failed to converge to a usable solution and gave poor accuracy (often equivalent to a blind guess, or 1 in 10 chance). Tables \ref{Tab3} and \ref{Tab4} give performance metrics that exclude ``outlying'' runs, that is, runs that gave less than mean accuracy of 0.5. The number of runs satisfying the mean accuracy requirement is given in the bottom row of \ref{Tab4}. This varied from 412 to 499 out of 500.

The algorithm ranking is not greatly affected by the outlier rejection, although the mean accuracy and mean IoU are. Removing outliers greatly reduces the standard deviation for both of the latter quantities and leads to a more confident ranking. Bearing in mind that all training was carried out for $K=10$ mixture components, one can summarise the detailed results as follows.
\begin{enumerate}
\item The best performance, both in mean accuracy and mean IoU, was obtained for the (2-D)x(3-D) and (3-D)x(3-D) versions of the SKEM. The mean accuracy of both was around 0.96.
\item The second best performance was obtained for 4-D, 5-D, 6-D and 9-D SKEM; all having mean accuracy around 0.90.
\item The third best performance, around 0.85 in mean accuracy, was obtained for the 3-D and 3x(1-D) SKEM versions, both of which do not use PCA features.
\item The fourth best performance, around 0.80 in mean accuracy, was obtained for the 13-D SKEM, which uses all the PCA and quadrant mass features.
\item The worst performance, around 0.71 in mean accuracy, was obtained for the 10-D SKEM, which is only based on PCA components.
\item All versions of the SKEM algorithm implemented, except for the 3D and 3x(1-D) versions, were quite sensitive to the initial choice of mean vector (for the same choice of initial covariance and mixture weights). The other 8 versions failed the 0.5 mean accuracy test in 10\% or more of the 500 runs.
\end{enumerate}

Table \ref{Tab6} gives a comparison of the mean accuracy and the maximum accuracy obtained on 500 runs for each of the 10 SKEM variants. Only runs satisfying the accuracy threshold are included in the average. The results show that the variants using only one type of feature, viz. the 3-D ones, which only use quadrant mass, and the 10-D, which only uses PCA features, have the lowest maximum accuracy on the test data set. All other variants attain 0.95 - 0.96 maximum accuracy for at least one of the initialisations in the 500 runs. As previously mentioned, the confusion matrices corresponding to the maximum accuracy runs appear in the appendix (section \ref{confmat}). The degree of ``generalisation,'' i.e. performance on other data sets, has not been assessed.

Plots of some of the estimated 6-D shared kernel models appear in Figs. \ref{fig6d1} and \ref{fig6d2}. Further plots appear in section \ref{plots} for both the 4-D and 6-D models. The particular graphical representation chosen for representing these multi-dimensional mixture PDFs is explained in that section. Generally speaking, for the digit classification problem considered here, ``unimodal'' class-conditioned mixture densities gave better performance than ``multimodal'' ones in the sense that the presence of a strongly multimodal class-conditioned model resulted in significant off-diagonal elements in the confusion matrix for the class in question. The reader is referred to sections \ref{plots} and \ref{confmat} for further details.

\begin{table}
\center
\begin{tabular}{|c|c|c|}
\hline
Alg & Mean Accuracy & Max Accuracy\\
\hline
\hline
3-D & 0.80 & 0.85 \\
\hline
3x(1-D) & 0.84 & 0.85 \\
\hline
4-D & 0.78 & 0.95 \\
\hline
5-D & 0.74 & 0.96 \\
\hline
(2-D)x(3-D) & 0.86 & 0.96 \\
\hline
6-D & 0.74 & 0.95 \\
\hline
(3-D)x(3-D) & 0.86 & 0.96 \\
\hline
9-D & 0.77 & 0.96 \\
\hline
10-D & 0.64 & 0.77 \\
\hline
13-D & 0.72 & 0.96 \\
\hline
\end{tabular}
\caption{Mean accuracy and maximum accuracy for different SKEM versions on 500 runs.}
\label{Tab6}
\end{table}

In summary, the following anecdotal conclusions can be drawn for this particular combination of training and test data for the digit data set.
\begin{enumerate}
\item Higher dimensional feature vectors seem to lead to worse performance when implemented as a single $M$-dimensional SKEM.
\item While low dimensional implementations are more robust to initialisation settings, they provide poor performance due to paucity of features.
\item SKEM classifiers using multiple feature types have better performance in terms of maximum achievable accuracy.
\item Better average accuracy performance (across initialisation settings) seems to be obtained by medium dimensional implementations.
\item The performance of medium dimensional implementations can be enhanced by decomposing the joint likelihood into a product of lower dimensional likelihoods obtained from non-overlapping subsets of features.
\end{enumerate}
The last point  is interesting because, although multiple features in the digit recognition problem are not independent, the assumption that they are group-wise independent leads to better performance. There may be a connection with mean-field theory approach to variational Bayesian optimisation (as described in \cite{Bishop2009}), which also decomposes a joint PDF (over the latent variables) into non-overlapping subsets.

\subsection{Variable Number of Mixture Components $K$}\label{Kvar}

The effect of different numbers of mixture components for the standard EM algorithm has been illustrated in \cite{Pulford2020b}, and we provide further experiments for the SKEM in this section. All results presented in section \ref{results} assumed $K=10$ mixture components. The choice of $K$ generally depends on the precise form of the data distribution being modelled. For standard EM, a data distribution with $K$ well-spaced discrete modes would require at least $K$ mixture components, while a highly non-Gaussian data distribution may require a larger number of components to accurately represent it. The situation for a shared kernel model in the multiple class case is less obvious. It seems reasonable to assume that at least $K=L$ components would be required for a shared kernel model with $L$ classes. This borne out in the simulations, where, regardless of the dimension of the feature vector, $K=L$ components gives reasonable performance for the digit recognition problem. The question is, therefore, what happens for values of $K$ less than or greater than $L$? To answer this question, we tested 4 versions of the SKEM algorithm on 100 runs for various values of $K\in\{3,6,8,10,12,15,18\}$:
\begin{enumerate}
\item 3-D: quadrant mass components 2:4 only (no PCA features);
\item (3+3)=6-D: quadrant mass with PCA components 2, 4 and 5;
\item (6+3)=9-D: quadrant mass with PCA components 2:7;
\item 10-D: PCA components 1:10 (with $\sigma=0.01$ dither on zero feature values).
\end{enumerate}

\begin{table}
\center
\begin{tabular}{|c|c|c|c|c|c|c|c|c|}
\hline
Alg & 3-D & Nr & 6-D & Nr & 9-D & Nr & 10-D & Nr\\
\hline
\hline
K=3 & 0.30$\pm 0.003$ & 100 & 0.30$\pm 0.003$ & 100 & 0.30$\pm 0.001$ & 100 & 0.30$\pm 0.002$ & 100\\
\hline
K=6 & 0.60$\pm 0.007$ & 100 & 0.60$\pm 0.014$ & 100 & 0.60$\pm 0.014$ & 96 & 0.56$\pm 0.01$ & 89\\
\hline
K=8 & 0.76$\pm 0.04$ & 99 & 0.78$\pm 0.043$ & 97 & 0.76$\pm 0.054$ & 95 & 0.69$\pm 0.05$ & 97\\
\hline
K=10 & 0.83$\pm 0.03$ & 97 & 0.86$\pm 0.06$ & 80 & 0.84$\pm 0.08$ & 83 & 0.69$\pm 0.05$ & 91\\
\hline
K=12 & 0.84$\pm 0.044$ & 88 & 0.88$\pm 0.08$ & 67 & 0.87$\pm 0.09$ & 73 & 0.71$\pm 0.043$ & 84\\
\hline
K=15 & 0.84$\pm 0.04$ & 77 & 0.83$\pm 0.15$ & 46 & 0.84$\pm 0.12$ & 43 & 0.71$\pm 0.06$ & 68\\
\hline
K=18 & 0.84$\pm 0.05$ & 61 & 0.86$\pm 0.17$ & 15 & 0.79$\pm 0.15$ & 22 & 0.73$\pm 0.07$ & 53\\
\hline
\end{tabular}
\caption{Mean accuracy (with standard deviation) of different SKEM versions on 100 runs. Nr is the number of runs satisfying the accuracy threshold, except for $K=3$ where all runs were below the threshold.}
\label{Tab5}
\end{table}

The results for mean accuracy are shown in Table \ref{Tab5}. For $K=3$ components, all tested versions of the SKEM algorithm had mean accuracy of 0.3. From the confusion matrices (not provided) it is clear that the classifier accurately recognised 3 of the 10 digits while consistently misclassifying all the other 7 digits. Similar behaviour was observed for $K=6$ components: all versions of the SKEM algorithm had mean accuracy of 0.6 except the 10-D version, whose mean accuracy was 0.56. In this case, the classifier accurately recognised 6 of the 10 digits while consistently misclassifying all the other 4 digits. Similarly, when $K=8$, the SKEM classifier can at best recognise 8 of the 10 digits, misclassifying the two others 100\% of the time. Thus when $K$ is less than the number of classes $L$, the performance of the classifier is limited because the number of assumed mixture components cannot adequately model the real data distribution for all the clases. The classifier can accuractely capture up to $K$ of the classes, with the other classes being incorrectly classified as one of the representable classes.

For $K\geq L=10$, performance is more nuanced. In all 4 SKEM variants, near optimal performance is achieved on the digits data set once $K=10$, with marginal improvement for $K=12$. For larger values of $K$, e.g. 15 and 18, a performance plateau in mean accuracy is reached. On the other hand, apart from requiring considerably more computation for larger values of $K$, the SKEM algorithm becomes much more sensitive to initialisation, frequently diverging for values of $K$ 50\% bigger than $L$, and tending to give a larger variation in mean accuracy. At $K=18$ the divergence problem becomes pronounced, with only 22\% or fewer of the 6-D and 9-D SKEM variants converging in 100 runs. Additionally, it is likely that the higher values of $K$ would lead to so-called ``overfit,'' which means that the extra degrees of freedom provided by the larger number of parameters allow the SKEM to model minor variations in the training data that are more ``noise-like'' than ``signal like.'' Although not assessed here, this typically leads to poor generalisation performance on test data that may differ substantially from the training data.

The conclusion is that there is an optimal range for the number of components $K$, regardless of the dimension of the feature vector. Choosing $K$ smaller than the number of classes results in poor performance due to lack of representational capacity in the shared kernel model. Large values of $K$, e.g. 50\% bigger than values in the optimal range, result in an increased tendency of the SKEM algorithm to diverge.

\section{Conclusions and Further Work}\label{conc}
The EM algorithm is a fundamental, unsupervised learning technique for estimating the parameters in a finite mixture representation of a multi-dimensional data set. Understanding how the EM algorithm works requires an understanding of the role of hidden or latent variables in simplifying the expression for the data likelihood. The price of this simplicity is that the estimator becomes iterative, repeating the E-step and M-step until convergence. The E-step is the computation of Baum's auxiliary function $Q(\Theta,\Theta_0)$. The M-step takes this explicit expression and computes updated mixture parameter estimates via a maximisation operation. The derivation of the standard EM algorithm relies on a vector of binary indicator variables. The derivation of EM that we presented uses $K$-valued scalar ``data association'' variables. This type of approach is common in the field of multiple target tracking. Our presentation, it is hoped, will help AI practitioners to gain a better appreciation of the role of latent variables in modern machine learning techniques that use the variational lower bound.

We then considered the more general shared kernel representation, which is a supervised learning approach requiring class labels. Previous treatments of this topic have failed to provide rigorous justification for the modified form of the EM algorithm in this case. Using class-conditioned indexing and applying data association techniques, we provided thorough derivations of (i) the complete data likelihood; (ii) Baum's auxiliary function (the E-step) and (iii) the maximisation (M-step) in the case of Gaussian shared kernel models. The subsequent algorithm, called shared kernel EM (SKEM), was then applied to a digit recognition problem using a novel 7-segment digit representation. Variants of the algorithm based on different numbers of features and different dimensions were compared in terms of mean accuracy and mean IoU. The effect of different numbers of assumed mixture components $K$ was also investigated. The results from these simulations indicate that: (i) higher dimensional feature vectors perform worse when implemented as a single $M$-dimensional SKEM; (ii) better average accuracy performance (across initialisation settings) seems to be obtained by medium dimensional implementations; (iii) the performance of medium dimensional implementations can be enhanced by decomposing the joint likelihood into a product of lower dimensional likelihoods obtained from non-overlapping subsets of features; (iv) there is an optimal range for the number of components $K$.

In terms of further work, it is recommended that the SKEM algorithm for classification should be compared with other supervised machine learning approaches such as deep convolutional neural networks on more varied data sets, starting with MNIST. Any comparison should consider the trade-off between performance and complexity, as well as the efficiency of the classifier in terms of its training data requirements. A better initialisation strategy is required than the random mean with equal covariances used in the numerical experiments presented here. Another interesting direction is the connection with the mean-field theory approach to variational Bayesian optimisation, which also decomposes a joint PDF (over the latent variables) into non-overlapping subsets of variables.

\bibliographystyle{unsrt}
\bibliography{skem}

\begin{thebibliography}{10}

\bibitem{Baum2}
{L. E. Baum, T. Petrie, G. Soules, and N. Weiss}.
\newblock {``A Maximization Technique Occurring in the Statistical Analysis of
  Probabilistic Functions of Markov Chains'', {\em Ann. Math. Statistics}, vol.
  41, pp. 164--171, 1970.}

\bibitem{Titterington}
{D. M. Titterington, A. F. M. Smith, and U. E. Makov}.
\newblock { {\em Statistical Analysis of Finite Mixture Distributions}, John
  Wiley, NY, 1985.}

\bibitem{Barshalom1}
{Y. Bar-Shalom, and E. Tse}.
\newblock {``Tracking in a Cluttered Environment With Probabilistic Data
  Association'', {\em Automatica}, vol. 11, pp. 451--460, 1975.}

\bibitem{Fortmann2}
{T. E. Fortmann, Y. Bar-Shalom, and M. Scheffe}.
\newblock {``Sonar Tracking of Multiple Targets Using Joint Probabilistic Data
  Association'', {\em IEEE J. Oceanic Eng.}, vol. OE-8, no. 3, pp. 173--184,
  July 1983.}

\bibitem{Luttrell}
{S. P. Luttrell}.
\newblock {``Partitioned Mixture Distribution: an Adaptive Bayesian Framework
  for Low-Level Image Processing'', {\em IEE Proc. Vis. Image Signal Process.},
  vol. 141, no. 4, pp. 251--260, Aug. 1994.}

\bibitem{Jarrad}
{G. Jarrad and D. W. McMichael}.
\newblock {``Shared Mixture Distributions and Shared Mixture Classifiers'',
  {\em Proc. Int. Conf. on Dec. and Control}, IDC'99, Adelaide, Feb. 1999.}

\bibitem{Titsias}
{M. K. Titsias and A. C. Likas}.
\newblock {``Shared kernel models for class conditional density estimation'',
  {\em IEEE Trans. Neural Networks}, vol. 12, no. 5, pp. 987--997, Sept. 2001.}

\bibitem{Titsias2003}
{M. K. Titsias and A. Likas}.
\newblock {``Class Conditional Density Estimation Using Mixtures with
  Constrained Component Sharing'', {\em IEEE Trans. PAMI}, vol. 25, no. 7, pp.
  924--928, July 2003.}

\bibitem{Dharanipragada}
{S. Dharanipragada and K. Visweswariah}.
\newblock {``Gaussian Mixture Models with Covariances or Precisions in Shared
  Multiple Subspaces'', {\em IEEE Trans. Audio, Speech \& Language Process.},
  vol. 14, no. 4, pp. 1255--1266, July, 2006.}

\bibitem{Neal99}
{R. Neal and G. Hinton}.
\newblock {``A view of the EM algorithm that justifies incremental, sparse, and
  other variants.'' In M. I. Jordan (ed.) {\em Learning in Graphical Models},
  MIT Press, Cambridge, MA, 1999.}

\bibitem{Bishop2009}
C.~Bishop.
\newblock {\em {Pattern Recognition and Machine Learning}}.
\newblock Springer, NY, 2009.

\bibitem{Pulford2020b}
{G. W. Pulford}.
\newblock {``From the Expectation Maximisation Algorithm to Autoencoded
  Variational Bayes'', October 2020, http://arxiv.org/abs/2010.13551.}

\bibitem{LeCun2}
{Y. LeCun, C. Cortes, and C. J. C. Burges}.
\newblock {The MNIST Database of Handwritten Digits. Available at:
  http://yann.lecun.com/exdb/mnist/}.

\bibitem{Dempster}
{A. P. Dempster, N. M. Laird, and D. B. Rubin}.
\newblock {``Maximum Likelihood from Incomplete Data via the {\em EM}
  Algorithm'', {\em J. Royal Statistical Soc.}, vol. 39, no. 1, pp. 1--38,
  1977.}

\bibitem{Bilmes}
{J. A. Bilmes}.
\newblock {``A Gentle Tutorial of the EM Algorithm and its Application to
  Parameter Estimation for Gaussian Mixture and Hidden Markov Models'', Dept.
  of Electrical Engineering and Computer Science, U.C. Berkeley, TR-97-021, pp.
  164--171, April 1998.}

\bibitem{Magnus}
{J. R. Magnus and H. Neudecker}.
\newblock {{\em Matrix Differential Calculus with Applications in Statistics
  and Econometrics}, John Wiley \& Sons, Great Britain, 1994.}

\bibitem{Petersen}
{K. R. Petersen and M. S. Pedersen}.
\newblock {``The Matrix Cookbook'', Nov. 2008. Available at:
  http://matrixcookbook.com}.

\bibitem{LeCun}
{Y. LeCun, L. Bottou, Y. Bengio, and P. Haffner}.
\newblock {``Gradient-based learning applied to document recognition'', {\em
  Proceedings of the IEEE}, vol. 86, no. 11, pp. 2278--2324, 1998.}

\bibitem{Keysers}
{D. Keysers, T. Deselaers, C. Gollan, and H. Ney}.
\newblock {``Deformation Models for Image Recognition'', {\em IEEE Trans.
  PAMI}, vol. 29, no. 8, pp. 1422--1435, Aug. 2007.}

\end{thebibliography}

\section{Confusion Matrices}\label{confmat}
Confusion matrices for the best (highest mean accuracy) run of each SKEM algorithm type are listed in this section. The indexing for rows and columns, in terms of digits, is 1 to 9, with 0 in the tenth position.

{\small
\[
C^*_{3{\rm D}}=\left[
\begin{array}{cccccccccc}
8 & 43 & 0 & 0 & 0 & 0 & 0 & 0 & 199 & 0\\
2 & 160 & 6 & 0 & 0 & 1 & 2 & 0 & 76 & 3\\
0 & 0 & 250 & 0 & 0 & 0 & 0 & 0 & 0 & 0\\
0 & 2 & 0 & 248 & 0 & 0 & 0 & 0 & 0 & 0\\
0 & 0 & 0 & 0 & 250 & 0 & 0 & 0 & 0 & 0\\
0 & 2 & 0 & 0 & 0 & 248 & 0 & 0 & 0 & 0\\
0 & 2 & 0 & 0 & 0 & 0 & 248 & 0 & 0 & 0\\
0 & 0 & 0 & 0 & 0 & 0 & 0 & 250 & 0 & 0\\
4 & 30 & 0 & 0 & 0 & 0 & 0 & 0 & 216 & 0\\
0 & 2 & 0 & 0 & 0 & 0 & 0 & 0 & 0 & 248
\end{array}
\right]
\]
}
{\small
\[
C^*_{3\times 1{\rm D}}=\left[
\begin{array}{cccccccccc}
101 & 32 & 0 & 0 & 0 & 0 & 0 & 0 & 116 & 1\\
49 & 128 & 12 & 2 & 0 & 3 & 4 & 0 & 48 & 4\\
0 & 0 & 250 & 0 & 0 & 0 & 0 & 0 & 0 & 0\\
0 & 0 & 0 & 250 & 0 & 0 & 0 & 0 & 0 & 0\\
0 & 0 & 0 & 0 & 249 & 0 & 0 & 0 & 0 & 1\\
0 & 2 & 0 & 0 & 0 & 247 & 0 & 0 & 0 & 1\\
0 & 8 & 0 & 0 & 0 & 0 & 242 & 0 & 0 & 0\\
0 & 0 & 0 & 0 & 0 & 0 & 0 & 250 & 0 & 0\\
73 & 20 & 0 & 0 & 0 & 0 & 1 & 0 & 156 & 0\\
0 & 4 & 0 & 0 & 0 & 0 & 0 & 0 & 0 & 246
\end{array}
\right]
\]
}
{\small
\[
C^*_{4{\rm D}}=\left[
\begin{array}{cccccccccc}
162 & 0 & 0 & 0 & 0 & 0 & 0 & 0 & 88 & 0\\
0 & 246 & 1 & 0 & 0 & 2 & 0 & 0 & 1 & 0\\
0 & 0 & 250 & 0 & 0 & 0 & 0 & 0 & 0 & 0\\
0 & 0 & 0 & 250 & 0 & 0 & 0 & 0 & 0 & 0\\
0 & 0 & 0 & 0 & 250 & 0 & 0 & 0 & 0 & 0\\
0 & 2 & 0 & 0 & 0 & 248 & 0 & 0 & 0 & 0\\
0 & 1 & 0 & 0 & 0 & 0 & 249 & 0 & 0 & 0\\
0 & 0 & 0 & 0 & 0 & 0 & 0 & 250 & 0 & 0\\
36 & 1 & 0 & 0 & 0 & 0 & 1 & 0 & 212 & 0\\
0 & 0 & 0 & 0 & 0 & 0 & 0 & 0 & 0 & 250
\end{array}
\right]
\]
}
{\small
\[
C^*_{5{\rm D}}=\left[
\begin{array}{cccccccccc}
169 & 0 & 0 & 0 & 0 & 0 & 0 & 0 & 81 & 0\\
0 & 250 & 0 & 0 & 0 & 0 & 0 & 0 & 0 & 0\\
0 & 0 & 250 & 0 & 0 & 0 & 0 & 0 & 0 & 0\\
0 & 0 & 0 & 250 & 0 & 0 & 0 & 0 & 0 & 0\\
0 & 0 & 0 & 0 & 250 & 0 & 0 & 0 & 0 & 0\\
0 & 0 & 0 & 0 & 0 & 250 & 0 & 0 & 0 & 0\\
0 & 0 & 0 & 0 & 0 & 0 & 250 & 0 & 0 & 0\\
0 & 0 & 0 & 0 & 0 & 0 & 0 & 250 & 0 & 0\\
29 & 0 & 0 & 0 & 0 & 0 & 1 & 0 & 220 & 0\\
0 & 0 & 0 & 0 & 0 & 0 & 0 & 0 & 0 & 250
\end{array}
\right]
\]
}
{\small
\[
C^*_{2{\rm D}\times 3{\rm D}}=\left[
\begin{array}{cccccccccc}
189 & 0 & 0 & 0 & 0 & 0 & 0 & 0 & 61 & 0\\
0 & 250 & 0 & 0 & 0 & 0 & 0 & 0 & 0 & 0\\
0 & 0 & 250 & 0 & 0 & 0 & 0 & 0 & 0 & 0\\
0 & 0 & 0 & 250 & 0 & 0 & 0 & 0 & 0 & 0\\
0 & 0 & 0 & 0 & 250 & 0 & 0 & 0 & 0 & 0\\
0 & 0 & 0 & 0 & 0 & 250 & 0 & 0 & 0 & 0\\
0 & 0 & 0 & 0 & 0 & 0 & 250 & 0 & 0 & 0\\
0 & 0 & 0 & 0 & 0 & 0 & 0 & 250 & 0 & 0\\
31 & 0 & 0 & 0 & 0 & 0 & 1 & 0 & 218 & 0\\
0 & 0 & 0 & 0 & 0 & 0 & 0 & 0 & 0 & 250
\end{array}
\right]
\]
}
{\small
\[
C^*_{6{\rm D}}=\left[
\begin{array}{cccccccccc}
166 & 0 & 0 & 0 & 0 & 0 & 0 & 0 & 84 & 0\\
0 & 250 & 0 & 0 & 0 & 0 & 0 & 0 & 0 & 0\\
0 & 0 & 250 & 0 & 0 & 0 & 0 & 0 & 0 & 0\\
0 & 0 & 0 & 250 & 0 & 0 & 0 & 0 & 0 & 0\\
0 & 0 & 0 & 0 & 250 & 0 & 0 & 0 & 0 & 0\\
0 & 0 & 0 & 0 & 0 & 250 & 0 & 0 & 0 & 0\\
0 & 0 & 0 & 0 & 0 & 1 & 249 & 0 & 0 & 0\\
0 & 0 & 0 & 0 & 0 & 0 & 0 & 250 & 0 & 0\\
27 & 0 & 0 & 0 & 0 & 0 & 1 & 0 & 222 & 0\\
0 & 0 & 0 & 0 & 0 & 0 & 0 & 0 & 0 & 250
\end{array}
\right]
\]
}
{\small
\[
C^*_{3{\rm D}\times 3{\rm D}}=\left[
\begin{array}{cccccccccc}
198 & 0 & 0 & 0 & 0 & 0 & 0 & 0 & 52 & 0\\
0 & 250 & 0 & 0 & 0 & 0 & 0 & 0 & 0 & 0\\
0 & 0 & 250 & 0 & 0 & 0 & 0 & 0 & 0 & 0\\
0 & 0 & 0 & 250 & 0 & 0 & 0 & 0 & 0 & 0\\
0 & 0 & 0 & 0 & 250 & 0 & 0 & 0 & 0 & 0\\
0 & 0 & 0 & 0 & 0 & 250 & 0 & 0 & 0 & 0\\
0 & 0 & 0 & 0 & 0 & 0 & 250 & 0 & 0 & 0\\
0 & 0 & 0 & 0 & 0 & 0 & 0 & 250 & 0 & 0\\
40 & 0 & 0 & 0 & 0 & 0 & 1 & 0 & 209 & 0\\
0 & 0 & 0 & 0 & 0 & 0 & 0 & 0 & 0 & 250
\end{array}
\right]
\]
}
{\small
\[
C^*_{9{\rm D}}=\left[
\begin{array}{cccccccccc}
179 & 0 & 0 & 0 & 0 & 0 & 0 & 0 & 71 & 0\\
0 & 250 & 0 & 0 & 0 & 0 & 0 & 0 & 0 & 0\\
0 & 0 & 250 & 0 & 0 & 0 & 0 & 0 & 0 & 0\\
0 & 0 & 0 & 250 & 0 & 0 & 0 & 0 & 0 & 0\\
0 & 0 & 0 & 0 & 250 & 0 & 0 & 0 & 0 & 0\\
0 & 0 & 0 & 0 & 0 & 249 & 0 & 0 & 0 & 1\\
0 & 0 & 0 & 0 & 0 & 0 & 250 & 0 & 0 & 0\\
0 & 0 & 0 & 0 & 0 & 0 & 0 & 250 & 0 & 0\\
32 & 0 & 0 & 0 & 0 & 0 & 0 & 0 & 218 & 0\\
0 & 0 & 0 & 0 & 0 & 0 & 0 & 0 & 0 & 250
\end{array}
\right]
\]
}
{\small
\[
C^*_{10{\rm D}}=\left[
\begin{array}{cccccccccc}
190 & 0 & 0 & 0 & 15 & 0 & 3 & 0 & 42 & 0\\
0 & 250 & 0 & 0 & 0 & 0 & 0 & 0 & 0 & 0\\
0 & 0 & 27 & 26 & 0 & 186 & 7 & 0 & 0 & 4\\
0 & 0 & 0 & 235 & 0 & 7 & 0 & 8 & 0 & 0\\
17 & 0 & 0 & 0 & 208 & 0 & 0 & 0 & 18 & 7\\
0 & 0 & 19 & 22 & 0 & 193 & 8 & 0 & 0 & 8\\
0 & 0 & 3 & 0 & 0 & 2 & 216 & 0 & 15 & 14\\
0 & 0 & 0 & 6 & 0 & 0 & 0 & 244 & 0 & 0\\
47 & 0 & 0 & 0 & 13 & 0 & 19 & 0 & 160 & 11\\
1 & 0 & 1 & 3 & 10 & 2 & 14 & 0 & 19 & 200
\end{array}
\right]
\]
}
{\small
\[
C^*_{13{\rm D}}=\left[
\begin{array}{cccccccccc}
201 & 0 & 0 & 0 & 0 & 0 & 0 & 0 & 49 & 0\\
0 & 250 & 0 & 0 & 0 & 0 & 0 & 0 & 0 & 0\\
0 & 0 & 250 & 0 & 0 & 0 & 0 & 0 & 0 & 0\\
0 & 0 & 0 & 250 & 0 & 0 & 0 & 0 & 0 & 0\\
0 & 0 & 0 & 0 & 250 & 0 & 0 & 0 & 0 & 0\\
0 & 0 & 0 & 0 & 0 & 250 & 0 & 0 & 0 & 0\\
0 & 0 & 0 & 0 & 0 & 1 & 249 & 0 & 0 & 0\\
0 & 0 & 0 & 0 & 0 & 0 & 0 & 250 & 0 & 0\\
52 & 0 & 0 & 0 & 0 & 0 & 0 & 0 & 198 & 0\\
0 & 0 & 0 & 0 & 0 & 0 & 0 & 0 & 0 & 250
\end{array}
\right]
\]
}

\newpage
\section{Shared Kernel Plots}\label{plots}

\begin{table}[b]
\center
\begin{tabular}{|c|c|c|c|c|c|c|c|}
\hline
Weight & $<$0.01 & 0.01 - 0.1 & 0.1 - 0.2 & 0.2 - 0.3 & 0.3 - 0.4 & 0.4 - 0.5 & $\geq$0.5\\
\hline
Thickness & 0.5 (dashed) & 0.5 & 1 & 2 & 3 & 4 & 5 \\
\hline
Colour & yellow & green & cyan & magenta & red & blue & black \\
\hline
\end{tabular}
\caption{Line thickness (in points) for ellipse plots as a function of mixture weight.}
\label{Tab7}
\end{table}

All of the SKEM algorithms simulated have feature vector dimension 3 or greater. It is therefore not possible to directly visualise the shared kernel models. A simple representation via 2-D projections of the $M$-D mixtures has been adopted here. We have selected 4-D and 6-D shared kernel models to illustrate some of the class-conditioned models that arose from training on the synthetic digit data. Importantly, in the case of a multivariate Gaussian PDF, the marginals can be obtained by extraction of the relevant components from the mean and covariance matrix, deleting the other variables. This means that for Gaussian mixtures, the 2-D projections are ``authentic'' in the sense that they are the true 2-D marginals of the $M$-D distribution, much as one can have a 2-D side-view of a 3-D object. In the 4-D mixture case, there are 3 possible 2-D projections $(x_1,x_2)$, $(x_1,x_3)$, $(x_2,x_3)$. In the 6-D mixture case, there are 15 such 2-D projections.

In each example, we selected one or more classes and plotted their shared kernel representations. The representation takes the form of a set of 1-sigma 2-D ellipses from the mixture model. Naturally these are the same for each class but the class-conditioned weights differ. The line thickness and colour was used to encode the weights according to Table \ref{Tab7}.

\begin{figure}[p]
\center
   \includegraphics[width=14cm]{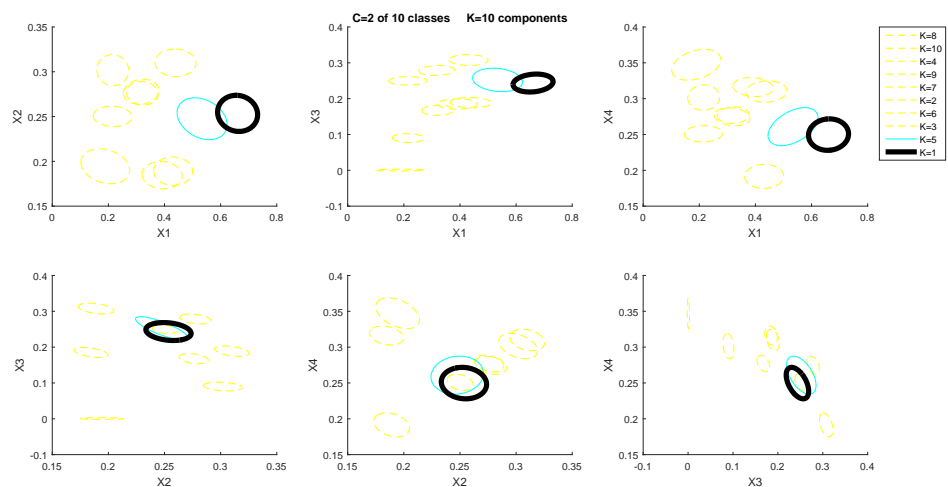}
\caption{2-D projections of 4-D shared kernel mixture for digit 2 on SKEM algorithm (3+1). Mixture weight encoded by line thickness as per Table \ref{Tab7}.}
\label{fig4d1}
\end{figure}

\begin{figure}[p]
\center
   \includegraphics[width=14cm]{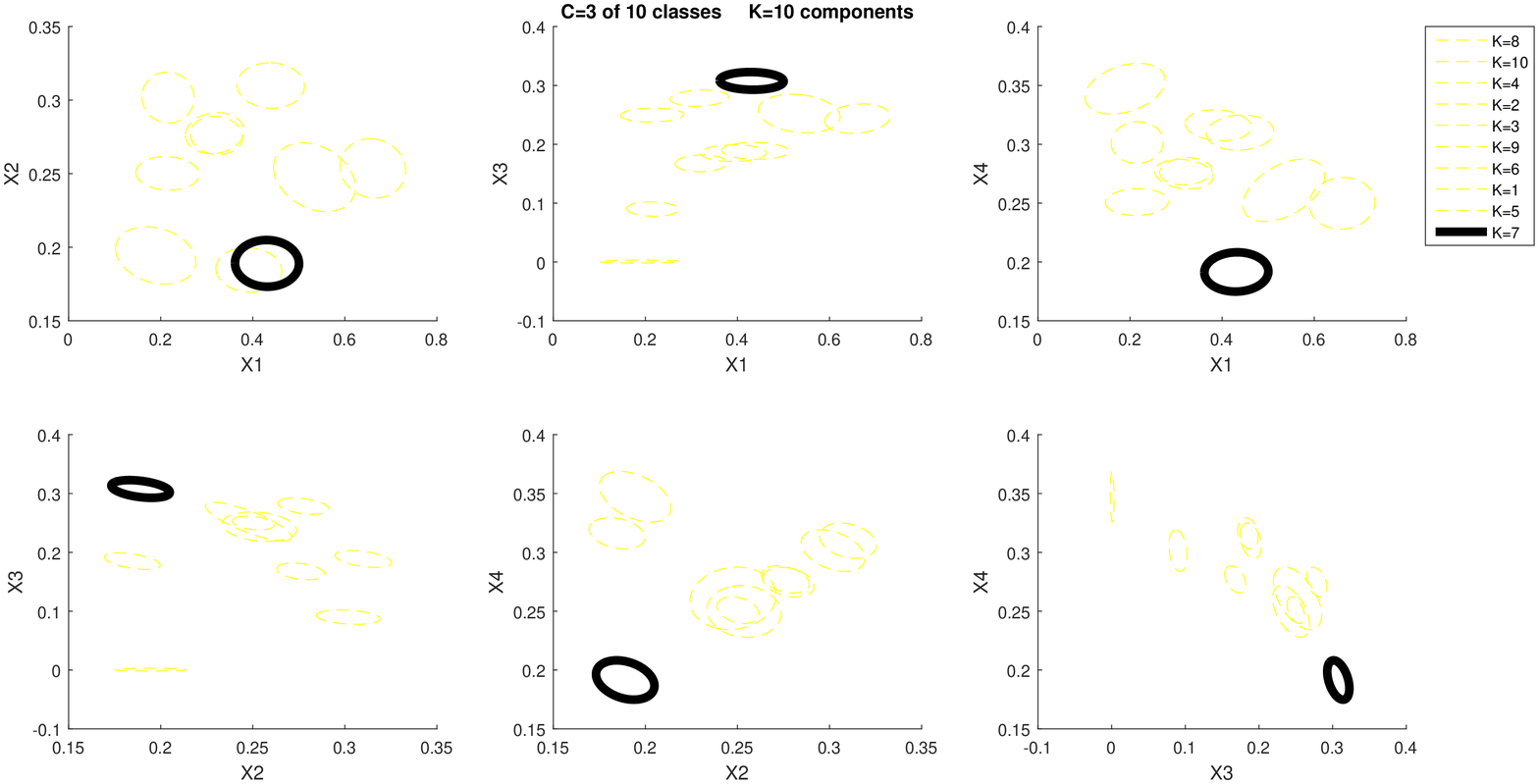}
\caption{2-D projections of 4-D shared kernel mixture for digit 3 on SKEM algorithm (3+1).}
\label{fig4d2}
\end{figure}

Figures \ref{fig4d1} and \ref{fig4d2} show 2-D projections of a shared kernel model for SKEM algorithm (3+1)=4-D for classes 2 and 3 respectively. This model had a mean accuracy of 0.9. These classes, corresponding to digits 2 and 3, had 245/250 and 250/250 correct classifications respectively. The class 2 model is ``bimodal'' whereas the class 3 model is ``unimodal.'' Here, unimodal means that there is one predominant mixture component, typically with weight 0.5 or greater, and multimodal means there are several (two or more) significant mixture components with weights 0.1 or greater. Visually this distinction shows up as a single black ellipse in the 2-D projections as opposed to a number of ellipses with solid outlines. Any mixture component with weight less than 0.01 has been shown as a dashed yellow ellipse in these plots.

\begin{figure}[p]
\center
   \includegraphics[width=14cm]{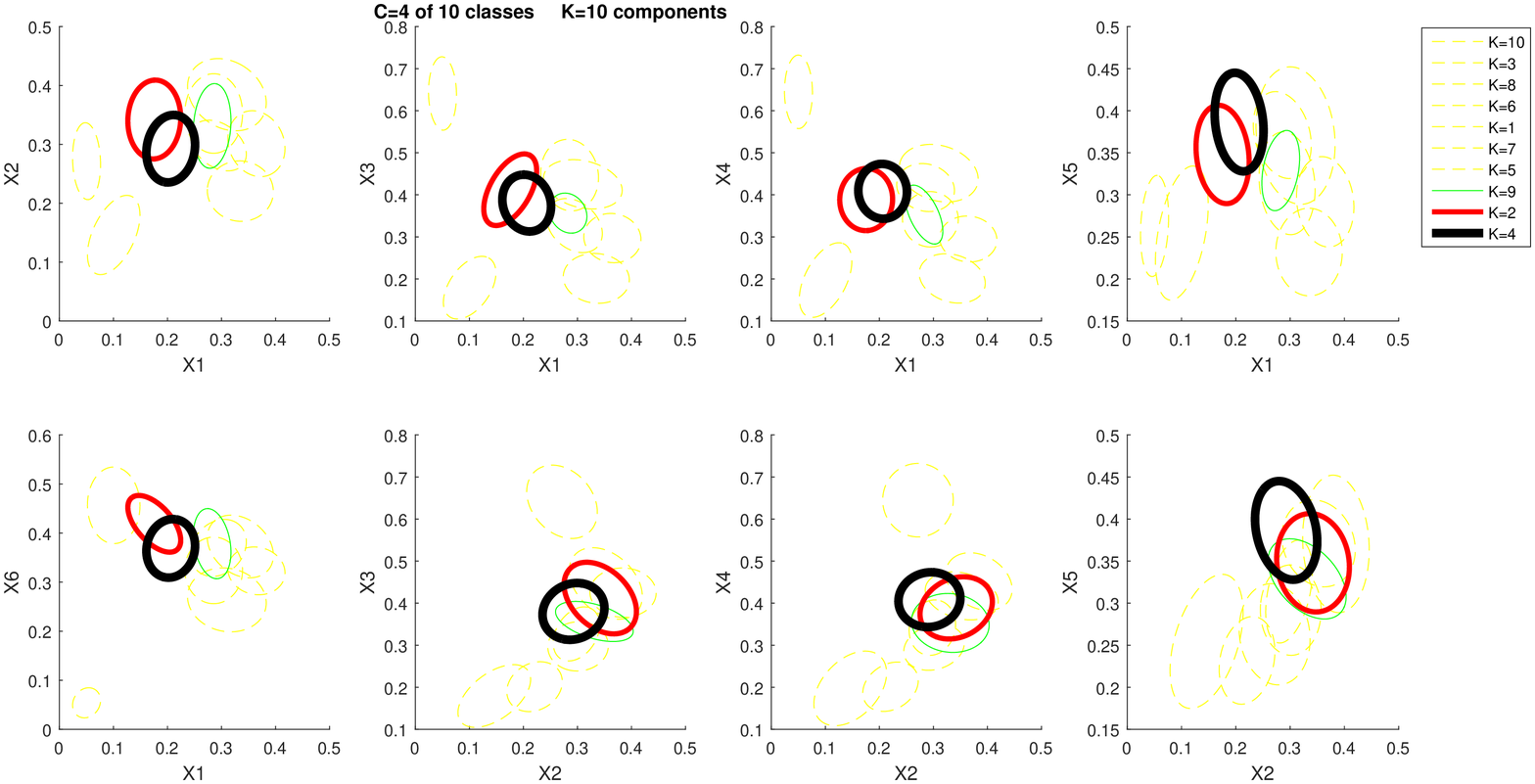}
\caption{2-D projections of 6-D shared kernel mixture for digit 4 on SKEM algorithm (3+1).}
\label{fig6d3}
\end{figure}

\begin{figure}[p]
\center
   \includegraphics[width=14cm]{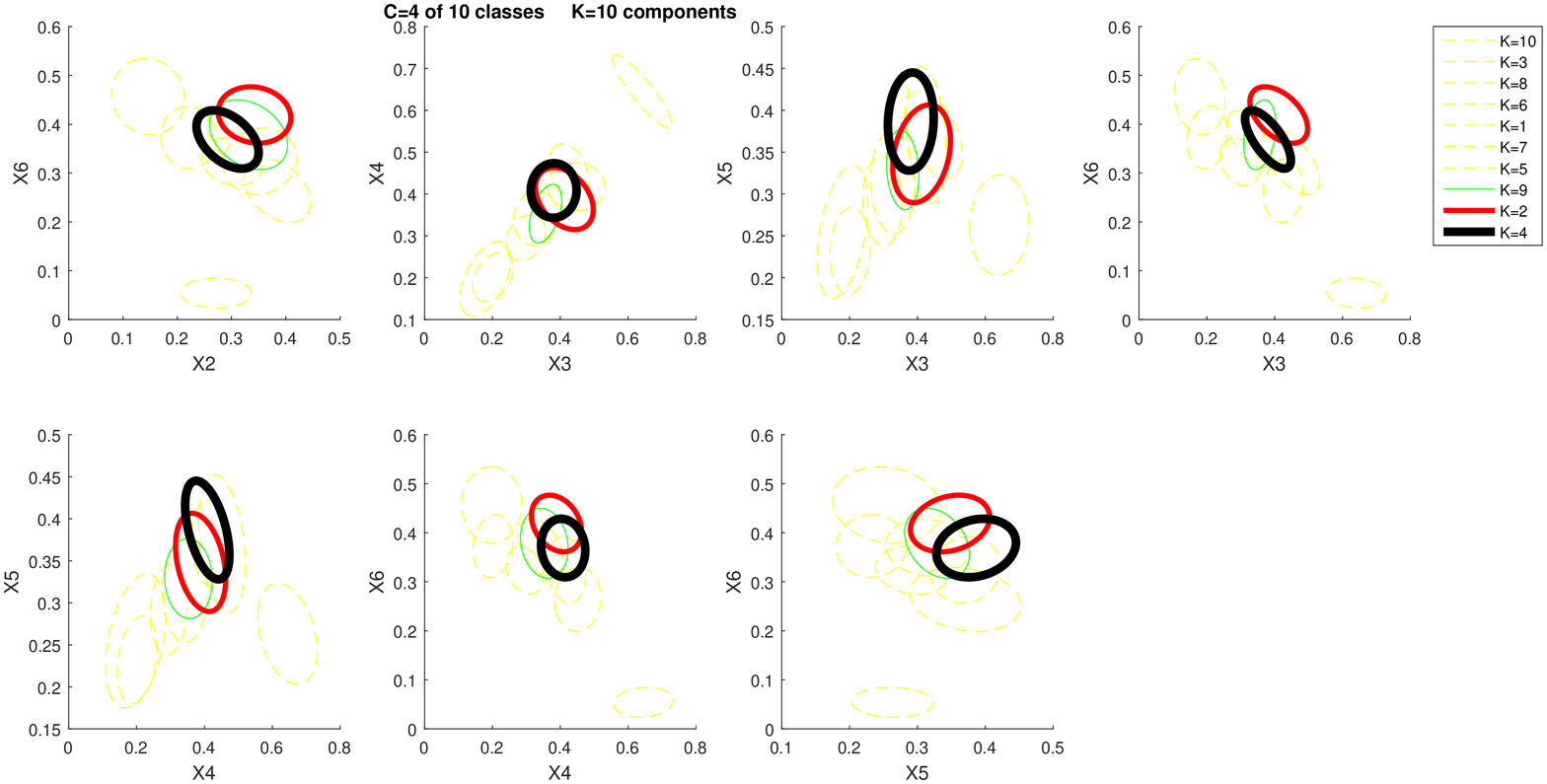}
\caption{2-D projections (continued) of 6-D shared kernel mixture for digit 4 on SKEM algorithm (3+1).}
\label{fig6d4}
\end{figure}

\begin{figure}[p]
\center
   \includegraphics[width=14cm]{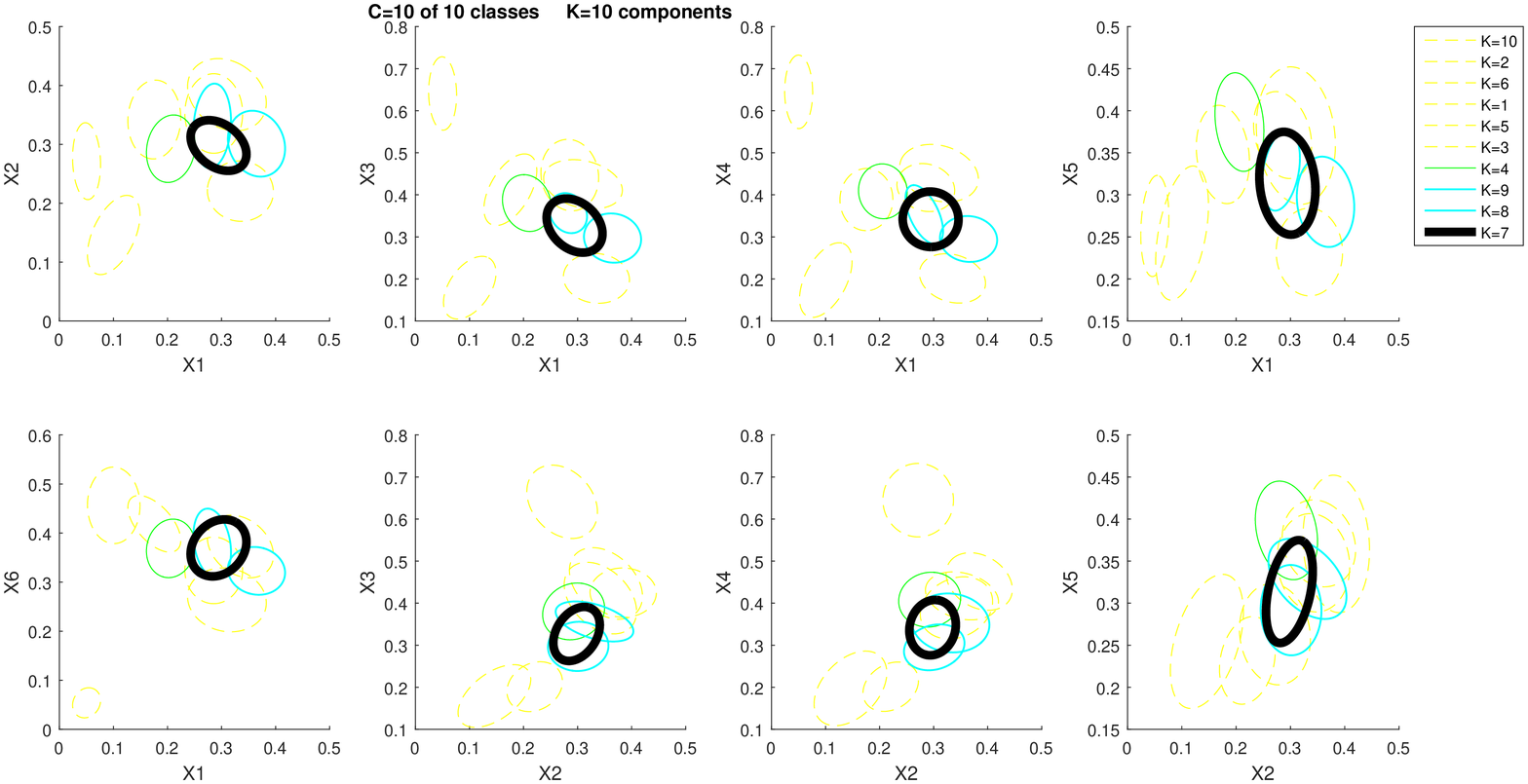}
\caption{2-D projections of 6-D shared kernel mixture for digit 0 on SKEM algorithm (3+1).}
\label{fig6d5}
\end{figure}

\begin{figure}[p]
\center
   \includegraphics[width=14cm]{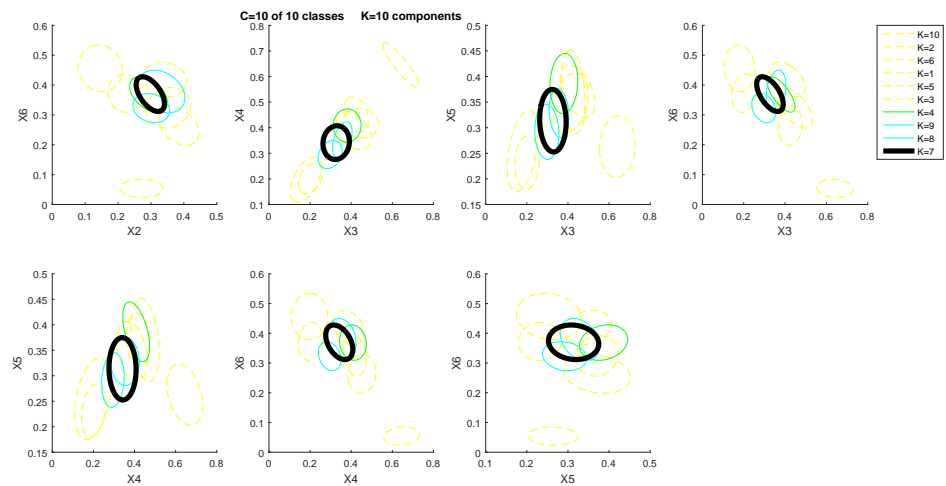}
\caption{2-D projections (continued) of 6-D shared kernel mixture for digit 0 on SKEM algorithm (3+1).}
\label{fig6d6}
\end{figure}

Figures \ref{fig6d3} -- \ref{fig6d6} show 2-D projections of a shared kernel model for SKEM algorithm (3+3)=6-D for classes 4 and 10 respectively, that is, for digits 4 and 0. The plots for digit 3 appear in Figs. \ref{fig6d1} and \ref{fig6d2}. While the average accuracy of this algorithm was around 0.74, this run's accuracy was only 0.56. The classification scores (diagonal entries of the confusion matrix) for this run were (for classes 3, 4 and 0): 97/250, 169/250 and 131/250. In all three cases, there were at least 3 significant mixture components and considerable multimodality.

The classification problem considered here has the characteristic that with 10 shared kernels, in many cases it is possible to accurately represent the observed feature vectors for a single digit with a single Gaussian kernel. This need not be the case in general. For instance, if the features coincide with the locations of the 7-segment display centres, the class-conditioned models would share many of the components; this would, however, imply prior knowledge of how the data were generated, and would probably increase the dimension of the feature vectors. The simulation also highlights the important role of feature extraction for machine learning methods that do not automatically extract features.

% Plots

\clearpage
\newpage
\section{Source Code}\label{src}
\subsection{7-Segment Digit Data Generation}\label{7seg}

\begin{figure}[hbt]
\center
   \includegraphics[height=7cm]{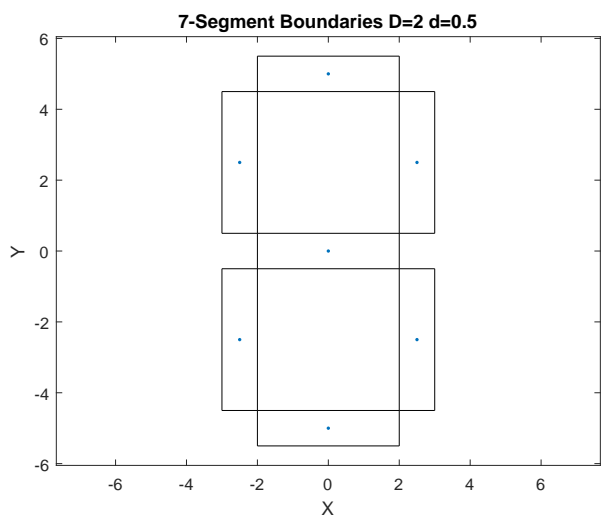}
\caption{Example of a 7-segment digit representation with $D=2,~d=0.5$.}
\label{fig7seg}
\end{figure}

The following code, compatible with Matlab and Octave, generates a set of noisy 28 x 28 greyscale images of the digits '0' to '9'. The resulting data set can be used for supervised training of the shared kernel EM algorithm. The choice of 28 x 28 greyscale images mimics LeCun's MNIST database of handwritten digit images. The digit representation is based on the so-called ``7-segment'' LED devices. The basic format, shown in Fig. \ref{fig7seg}, implemented by the function fig8\_gen(), is the well known figure-8 arrangement with each of the 7 segments being vertically or horizontally aligned. The code does not allow for changes in orientation (tilt) of the segments.

\begin{figure}
\center
\begin{tabular}{cc}
\includegraphics[width=6cm]{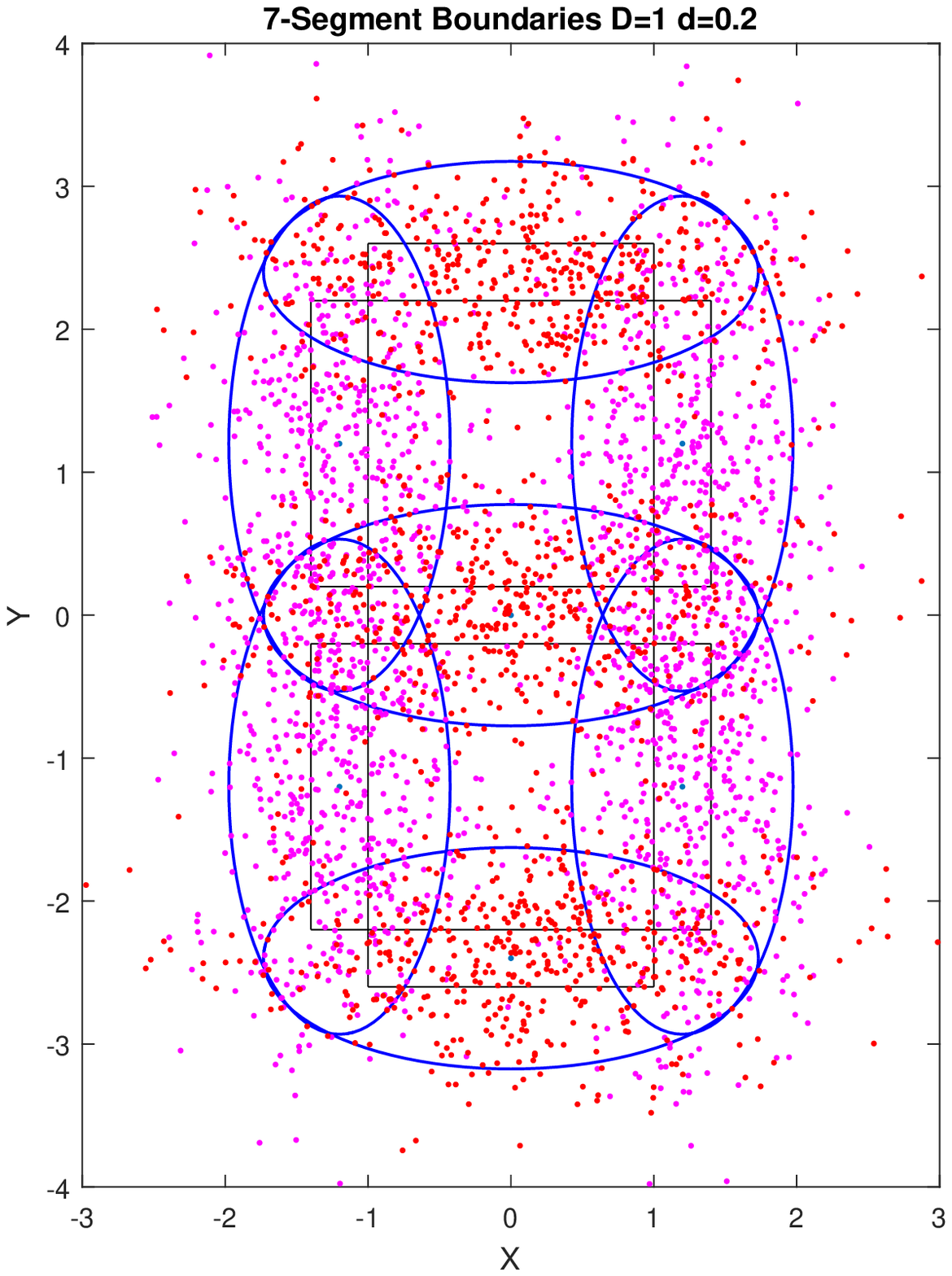} &
\includegraphics[width=6cm]{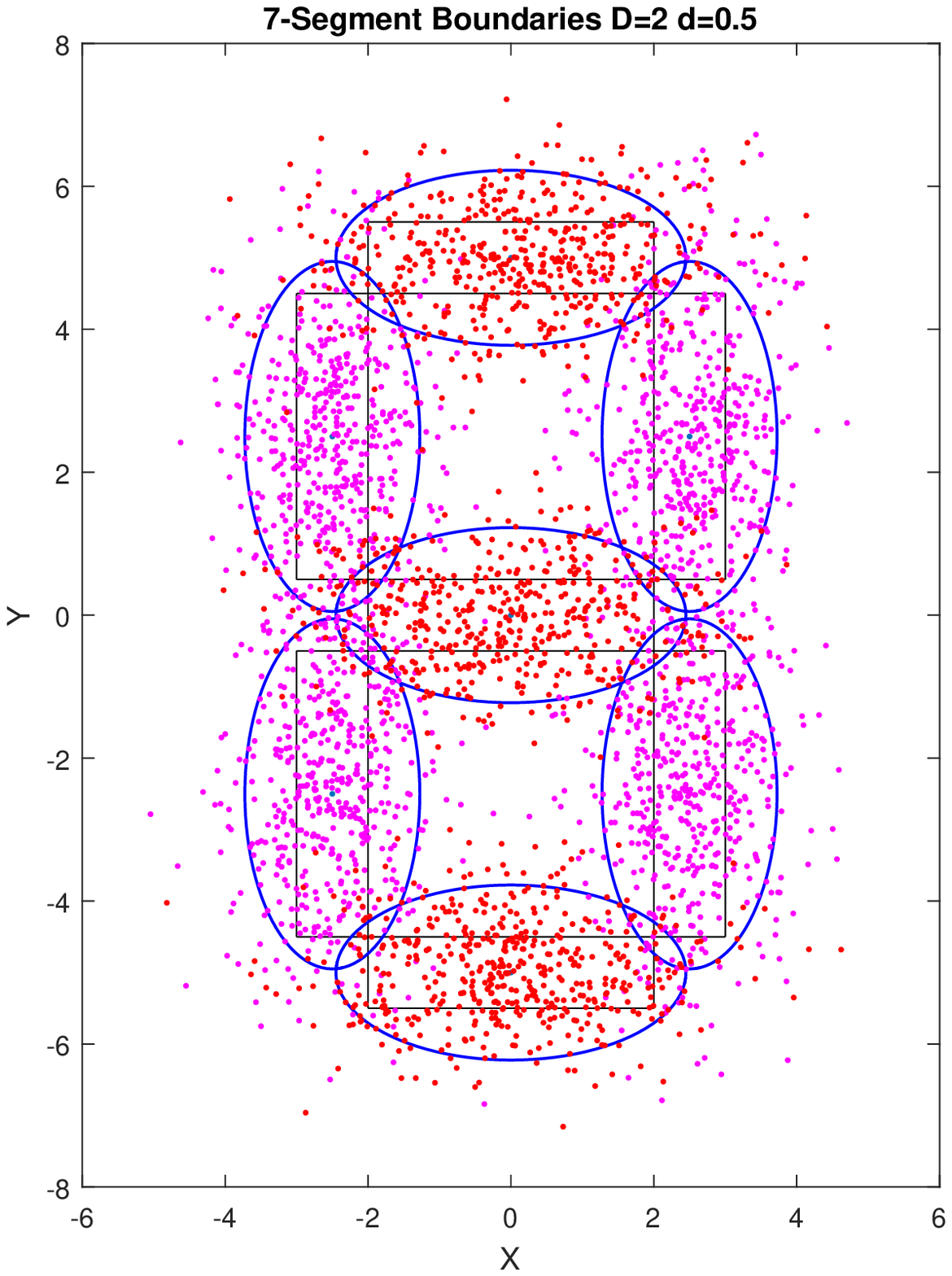}\\
\end{tabular}
\caption{7-segment display digits with $(D=1,~d=0.2)$ (left) and $(D=2,~d=0.5)$ (right), $3\sigma$ covariances ellipses, 500 random samples.}
\label{fig7seg2}
\end{figure}

The function digit\_data\_gen() works as follows. To generate noisy digit data, a Gaussian PDF is placed at the centre of each segment with the semi-axes of its covariance ellipse adjusted to equal $D$ and $d$, both multiplied by a factor of $n_{\sigma}$ to ensure good coverage of the segment under random sampling (see example in Fig. \ref{fig7seg2}). A bounding box with sides 1.8 times the actual size of the 7-segment display is defined as the image region. This region is divided into a grid of 28 x 28 cells (pixels). For each digit, the random samples from the Gaussian PDFs centred on the segments allocated to that digit are binned into the 28 x 28 cells. The pixel counts in each cell are scaled so that the maximum count in any one cell is 255, resulting in an 8-bit greyscale image of  the noisy digit. Finally, a database consisting of an equal number of each digit is generated using the function gen\_digit\_dataset(). Example calls are given in the header of the source code.

\subsection*{fig8\_gen.m}
\begin{verbatim}
% function [S,segcen,BB]=fig8_gen(D,d,Nsamples,N,[Nsig])
% Generate Nsamples samples for a digit N between 0 and 9
% from a "7-segment" based mixture.
% Nsig = 1 unless specified for covariance std scaling
%
% Segment size is (2*D) x (2*d)
% half width of display (between centres): w = D + d
% half height of display (between centres): h = 2*w
% Centre ordering (7 x 2 segcen matrix)
%    6
% 5     1
%    7
% 4     2
%    3
%
% Corner ordering (per 4 rows of 28 x 2 segcnr matrix)
% 1 top right
% 2 bottom right
% 3 top left
% 4 bottom left
% Bounding box: BB
%
% Example
% [S,segcen,BB]=fig8_gen(1,.1,100,6);
% plot(S(1,:),S(2,:),'.'); axis equal
%
% Copyright 2020 Graham Pulford

function [S,segcen,BB]=fig8_gen(D,d,Nsamples,N,varargin)
if nargin>4
    Nsig = varargin{1};
else
    Nsig=1;
end

w=D+d;
h=2*w;
% 7 x 2 matrix of (x,y) coords of segment centres
segcen=[w w; w -w; 0 -h ;-w -w; -w w; 0 h; 0 0];
BBx1=-(D+2*d);
BBx2=D+2*d;
BBy1=-(2*D+3*d);
BBy2=2*D+3*d;
BB=[BBx1,BBx2,BBy1,BBy2];

Px=Nsig*diag([D,d]);
Py=Nsig*diag([d,D]);

if N==0
    segv=[1,2,4,5];
    segh=[3,6];
elseif N==1
    segv=[1,2];
    segh=[];
    segcen(1,:)=[0 w]; % centre the 1 on the y-axis
    segcen(2,:)=[0 -w];
elseif N==2
    segv=[1,4];
    segh=[3,6,7];
elseif N==3
    segv=[1,2];
    segh=[3,6,7];
elseif N==4
    segv=[1,2,5];
    segh=[7];
elseif N==5
    segv=[2,5];
    segh=[3,6,7];
elseif N==6
    segv=[2,4,5];
    segh=[3,6,7];
elseif N==7
    segv=[1,2];
    segh=[6];
elseif N==8
    segv=[1,2,4,5];
    segh=[3,6,7];
elseif N==9
    segv=[1,2,5];
    segh=[3,6,7];
else
    error('unsupported digit. N must be between 0 and 9')
end

% Generate samples
S=[];
for i=1:length(segv)
    j=segv(i);
    g=gaussian_mu_P_multiple(segcen(j,:),Py,Nsamples);
    S=[S,g];
end
for i=1:length(segh)
    j=segh(i);
    g=gaussian_mu_P_multiple(segcen(j,:),Px,Nsamples);
    S=[S,g];
end
return
\end{verbatim}

\subsection*{gaussian\_mu\_P\_multiple.m}
\begin{verbatim}
function g = gaussian_mu_P_multiple(mu,P,Nsamples)
% Generate Nsamples random samples from a multivariate Gaussian PDF N(x;mu,P)
% with mean vector mu and covariance matrix P
n=length(mu);
mu=reshape(mu,n,1);
P=0.5*(P+P'); % ensure symmetry
T=transpose(chol(P)); % Chloesky factorisation of cov matrix such that P = T*T'
g=randn(length(mu),Nsamples); % Generate random sample from N(0, I)
g = T*g + repmat(mu,1,Nsamples); % Affine transformation to N(mu,P)
return
\end{verbatim}

\subsection*{digit\_data\_gen.m}
\begin{verbatim}
function Image = digit_data_gen(D,d,Nsamples,N,varargin)
% Generate a pixellated 28 x 28 greyscale image of a single digit
% using a 7-segment Gaussian mixture representation.
% Segment size is (2*D) x (2*d)
% D - half length of segment
% d - half width of segment
% Nsamples - number of samples for Gaussian on each segment
% N digit between 0 and 9
% Nsig = 1 unless specified for covariance std scaling
% do_plot - set to 1 for plot (0 = no plot)
% Image - 28 x 28 8-bit greyscale image output
%
% Example
% Image = digit_data_gen(2,0.5,100,6,1.2); imshow(Image,[])
%
% Copyright 2020 Graham Pulford

if nargin==4
    Nsig=1;
    do_plot=1;
elseif nargin==5
    Nsig=varargin{1};
    do_plot=1;
elseif nargin==6
    Nsig=varargin{1};
    do_plot = varargin{2};
end
BB_factor=1.8; % Allow for digit noise in BB definition
Ncells=28; % size of image is Ncells x Ncells
[S,segcen,BB]=fig8_gen(D,d,Nsamples,N,Nsig);
BB_big=BB_factor*[BB(3),BB(4),BB(3),BB(4)];
if do_plot
    figure(2)
    plot(S(1,:),S(2,:),'.'); axis(BB_big);
end
% limit all samples to the enlarged bounding box BB_big
S(1,S(1,:)>BB_big(4))=BB_big(4);
S(2,S(2,:)>BB_big(4))=BB_big(4);
S(1,S(1,:)<BB_big(3))=BB_big(3);
S(2,S(2,:)<BB_big(3))=BB_big(3);
dx=2*abs(BB_big(4))/Ncells; % grid cell size (square)
Image=zeros(Ncells);
for i=1:Ncells
    x1=BB_big(1)+(i-1)*dx;
    x2=x1+dx;
    I=intersect(find(S(1,:)>=x1),find(S(1,:)<x2));
    if ~isempty(I)
        for r=1:length(I)
            j=ceil((S(2,I(r))-BB_big(3))/dx);
            if j<=0
                j=1;
            end
            if j>28
                j=28;
            end
            Image(i,j)=Image(i,j)+1;
        end
    end
end
Imax=max(max(Image));
Image=(255/Imax).*Image; % scale to [0 - 255]
% image coordinate origin top left, vertical axis counts down
Image=flipud(Image');
Image=uint8(Image); % convert to unsigned 8-bit integer
if do_plot
    figure(3)
    imshow(Image,[])
end
\end{verbatim}

\subsection*{gen\_digit\_dataset.m}
\begin{verbatim}
function Images = gen_digit_dataset(D,d,Ngen,Nimages_per_digit,varargin)
% Generate a training database of pixellated 28 x 28 greyscale image
% of single digits from 0 - 9 using a "7-segment" Gaussian mixture representation.
% Segment size is (2*D) x (2*d)
% D - half length of segment
% d - half width of segment
% Ngen - number of samples for Gaussian on each segment
% Nimages_per_digit - number of 28 x 28 sample images for each digit
% Nsig = 1 unless specified for covariance std scaling
% Images - 4D array (10 x Nimages_per_digit x 28 x 28) of 8-bit greyscale images
%
% Copyright 2020 Graham Pulford

if nargin>4
    Nsig = varargin{1};
else
    Nsig=1;
end
Ncells=28;
digits=[1:9,0];
Images=zeros(10,Nimages_per_digit,Ncells,Ncells);
for i=1:10
    digit=digits(i);
    disp(['digit: ',num2str(digit)])
    for n=1:Nimages_per_digit
        if mod(n,100)==0
            disp(num2str(n))
        end
        Images(i,n,:,:)=digit_data_gen(D,d,Ngen,digit,Nsig,0);
    end
end
Images=uint8(Images);
return
\end{verbatim}

\subsection{Feature Extraction}\label{fextract}

\subsection*{calc\_pca\_features.m}
\begin{verbatim}
function [features,truth]=calc_pca_features(Images)
% Calculate features for digit dataset using PCA on 8-bit greyscale image.
% Most significant PCA components (row 1) down-selected to elements 11:20 of 28.
% Four quadrant relative mass measures (Q1, Q2, Q3, Q4) also included.
%
% Input
% Images - 4D array (10 x Nimages x 28 x 28) of 8-bit greyscale images
%
% Requires Matlab Statistics toolbox (R2015) function pca.m
% which normalises the data using (X-mu)/sigma computed on entire image.
%
% Example
% Images=gen_digit_dataset(2,0.5,100,500,1.2);
% [features,truth]=calc_pca_features(Images);
%
% Copyright 2020 Graham Pulford

[Ndigit,Nimg,Ncells1,Ncells2]=size(Images);
if Ndigit~=10
    error('Images dim 1 should be 10 digits')
end
if Ncells1~=28 || Ncells2~=28
    error('Images dims 3 & 4 should be 28')
end
Ncol1=11; % index of start column of Xp
Ncols=10; % number of columns of Xp to retain
Nvec=[1:9,0]; % actual digit
features=[];
truth=[];
for n=1:Nimg % digit sample
    for j=1:length(Nvec)
        N=Nvec(j); % index into Images array
        if N==0
            N=10;
        end
        X=double(squeeze(Images(N,n,:,:)));
        [q1,q2,q3,q4]=quadrant_mass(X);
        Xp=transpose(pca(X));
        Xp=Xp(1,Ncol1:Ncol1+Ncols-1); % only retain 1st column of PCA
        features=[features; Xp,q1,q2,q3,q4];
        truth=[truth;Nvec(j)];
    end
end
return
\end{verbatim}

\subsection*{quadrant\_mass.m}
\begin{verbatim}
function [q1,q2,q3,q4]=quadrant_mass(X)
% Compute relative mass in each quadrant of 2D grid
% quadrants indexed by 1. NE, 2, NW, 3. SW, 4. SE
%   Image
%   2  1
%   3  4
% NB indexing is array based, not image based, so x, y are swapped

X(X>0)=1; % binarise X for mass calculation
[m,n]=size(X);
m2=floor(m/2);
n2=floor(n/2);
Xtotal=sum(sum(X));
q1=sum(sum(X(1:m2,n2+1:n)))/Xtotal;
q2=sum(sum(X(1:m2,1:n2)))/Xtotal;
q3=sum(sum(X(m2+1:m,1:n2)))/Xtotal;
q4=sum(sum(X(m2+1:m,n2+1:n)))/Xtotal;
return
\end{verbatim}

\subsection{Source Code for Shared Kernel EM Algorithm}\label{srcEM}
The Matlab source code for the shared kernel EM algorithm is included in this appendix. It is compatible with any version since Matlab$^{\rm TM}$ 6 or Octave 4. The convergence test is hard coded for a tolerance of 0.1 on each class-conditioned log likelihood. The code is deliberately not vectorised so that it can be recoded easily in a faster language if desired. A simple data generation program is also provided that can generate $M$-dimensional samples from an arbitrary shared kernel Gaussian mixture model with $K$ components and $L=NC$ classes.

As an illustration, a $M=2$-dimensional example is provided. First generate $N=2000$ samples of simulated data $y$ using the function gen\_mixture():
\begin{verbatim}
Pi_true = [0.1,0.8,0.1; 0.7,0.1,0.2; 0.3,0.1,0.6];
Mu_true = [0,3,6; 2,1,3];
P_true = [0.5*eye(2), 0.5*eye(2), 0.5*eye(2)];
y = gen_mixture(Mu_true, P_true, Pi_true, 2000);
\end{verbatim}
Create the data matrix $X$ and the class label vector $C$:
\begin{verbatim}
X=[squeeze(y(1,:,:)) squeeze(y(2,:,:)) squeeze(y(3,:,:))];
C=[ones(1,2000),2*ones(1,2000),3*ones(1,2000)];
\end{verbatim}
Run the shared kernel EM algorithm by calling the function SKEM(). The algorithm will terminate when either the number of passes NP is reached or the convergence criterion is satisfied. Note that the code transposes the input data matrix to $N\times M$, if required, where $N>M$.
\begin{verbatim}
Pi0=1/3*ones(3);
Mu0=[-1,2,7; 0,1,2];
P0=[2*eye(2), 2*eye(2), 2*eye(2)];
[Pi,Mu,P] = SKEM(X, C, Mu0, P0, Pi0, 50)
\end{verbatim}
If correct convergence is achieved, which should be the case when the input data matches the model assumed in the shared kernel EM algorithm, the mixture means, covariances and weights at convergence should be close to their true values.

\subsection*{SKEM.m}
\begin{verbatim}
function [Pi,Mu,P] = SKEM(X,C,Mu,P,Pi,NP)
% EM algorithm for supervised learning of K-component NC-class shared kernel
% Gaussian mixture from N x M-dimensional observations
%
% Input
% X - M-D Data matrix (N x M) N=number of samples, M=number of features (dimension)
% C - vector of classes (same length as X)
% Mu - initial estimates of mixture means (M x K)
% P - initial estimates of mixture covariance matrices (M x KM) (variance if M=1)
% Pi - initial estimates of mixing probabilities (K x NC)
% NP - number of passes
%
% Output
% Mu - mixture means at termination (M x K)
% P - mixture covariance matrices at termination (M x KM)
% Pi - mixing probabilities at termination (K x NC)
%
% Copyright 2007 - 2020 Graham Pulford

TOL=1.0E-1; % tolerance for convergence criterion

[N,M]=size(X); % M = dimension, N = number of samples
if M>N
    disp('transposing data matrix to N x M with N > M')
    X=transpose(X);
    [N,M]=size(X);
end
if length(C)~=N
    error('data X and classes C must be same length')
end

classes=unique(C); % set of classes
NC=length(classes); % number of classes
if size(Pi,2)~=NC
    error('pi must have NC columns')
end
nij=cell(1,NC); % class-conditioned data set indexes
for c=1:NC
    nij{c}=find(C==classes(c));
end
Ni=zeros(1,NC);
for c=1:NC
    Ni(c)=length(nij{c});
end

K=size(Mu,2);

% allocate loop variables
G=zeros(K,N); % gaussian evaluations at X(n) for parameters Theta_k
W=zeros(K,N); % posterior probabilities
Sum_Log_Lik_prev=-Inf*ones(1,NC); % sum of data likelihoods (per class)
DLogLik=zeros(1,NC);
converged=zeros(1,NC);
for ipass=1:NP % EM loop
    disp(['pass ', num2str(ipass)]);
    for k=1:K
        k1=(k-1)*M+1;
        k2=k*M;
        for n=1:N
            G(k,n)=gaussian2(transpose(X(n,:))-Mu(:,k),P(:,k1:k2));
        end
    end
    disp(['MaxAbsMatrix(G): ', num2str(max(max(abs(G))))])

    % posterior weights N x K (measurement-to-component association probabilities)
    Sum_Log_Lik=zeros(1,NC);
    for c=1:NC
        for j=1:Ni(c)
            n=nij{c}(j);
            for k=1:K
               W(k,n)=Pi(k,c)*G(k,n);
            end
            sumk=sum(W(:,n)); % sum over k
            W(:,n)=W(:,n)./sumk;
            Sum_Log_Lik(c)=Sum_Log_Lik(c)+log(sumk);
        end
    end

    % mixture weights NC x K (class-to-component association probabilities)
    for c=1:NC
        for k=1:K
            Pi(k,c)=0;
            for j=1:Ni(c)
                n=nij{c}(j);
                Pi(k,c)=Pi(k,c)+W(k,n);
            end
        end
        Pi(:,c)=Pi(:,c)./Ni(c);
    end

    % mixture component means
    Mu=zeros(M,K);
    for k=1:K
        sumcj=0; % sum over c and j
        for c=1:NC
            for j=1:Ni(c)
                n=nij{c}(j);
                sumcj=sumcj+W(k,n);
                Mu(:,k)=Mu(:,k)+W(k,n)*transpose(X(n,:));
            end
        end
        Mu(:,k)=Mu(:,k)./sumcj;
    end

    % mixture component covariances
    for k=1:K
        k1=(k-1)*M+1;
        k2=k*M;
        P(:,k1:k2)=0;
        sumcj=0; % sum over c and j
        for c=1:NC
            for j=1:Ni(c)
                n=nij{c}(j);
                sumcj=sumcj+W(k,n);
                P(:,k1:k2)=P(:,k1:k2)+W(k,n)*(X(n,:)'-Mu(:,k))...
                           *transpose(X(n,:)'-Mu(:,k));
            end
        end
        P(:,k1:k2)=P(:,k1:k2)./sumcj;
    end

    for c=1:NC
        DLogLik(c)=Sum_Log_Lik(c)-Sum_Log_Lik_prev(c);
        if ipass>1
            if abs(DLogLik(c))<TOL % Log likelihood difference is < TOL
                converged(c)=1;
            end
        end
    end
    Sum_Log_Lik_prev=Sum_Log_Lik;

    if all(converged)
        disp(['Convergence at ipass: ',num2str(ipass)])
        break
    end
end
return
\end{verbatim}

\subsection*{gen\_mixture.m}
\begin{verbatim}
function y = gen_mixture(mu,P,Pi,N)
% Generate random samples from M-D shared kernel
% gaussian mixture density with NC classes
%
% Input
% mu - mean vectors for K components M x K
% P - covariance matrices for K components M x M x K (variance for 1-D case)
% Pi - mixture weights for each class NC x K
% N - number of data samples
%
% Output
% y - Data matrix NC x M x N
%
% Examples
% 1-D 3-class
% y = gen_mixture([0,3,6],[0.5,0.5,0.5],[0.1,0.8,0.1; 0.7,0.1,0.2; 0.3,0.1,0.6],2000);
% 2-D 3-class
% y = gen_mixture([0,3,6; 2,1,3],[0.5*eye(2),0.5*eye(2),0.5*eye(2)],...
%                 [0.1,0.8,0.1; 0.7,0.1,0.2; 0.3,0.1,0.6],2000);
%
% Copyright 2007 - 2020 Graham Pulford

[Dmu1,Dmu2]=size(mu); % M x K
[Dpi1,Dpi2]=size(Pi); % NC x K
[DP1,DP2]=size(P); % M x (MK)
if DP1>1 % M>1
    if Dmu1~=DP1 || Dmu2~=Dpi2
        error('Incompatible matrix sizes (M-D case): mu (MxK) P (MxMK) Pi (NCxK)')
    end
    if DP2/Dmu2~=Dmu1
        error('Incompatible matrix sizes (M-D case): P (MxMK)')
    end
else % M=1
    if Dmu1~=DP1 || Dmu2~=DP2 || Dmu2~=Dpi2
        error('Incompatible matrix sizes (1-D case): mu (1xK) P (1xK) Pi (NCxK)')
    end
end

M=Dmu1; % dimension of data
K=Dmu2; % number of components
NC=Dpi1; % number of classes

if M==1 % 1-D case
    y=zeros(NC,N);
    for c=1:NC
        for i=1:N
            u=rand;
            % determine the component that is selected
            for k=1:K
                if sum(Pi(c,1:k))>u
                    j=k;
                    break
                end
            end
            y(c,i)=gaussian_mu_P(mu(j),P(j));
        end
    end
else % M-D case, M>1
    y=zeros(NC,M,N);
    for c=1:NC
        for i=1:N
            u=rand;
            % determine the component that is selected
            for k=1:K
                if sum(Pi(c,1:k))>u
                    j=k;
                    break
                end
            end
            y(c,:,i)=gaussian_mu_P(mu(:,j),P(:,(j-1)*M+1:j*M));
        end
    end
end
return
\end{verbatim}

\subsection*{gaussian\_mu\_P.m}
\begin{verbatim}
function g = gaussian_mu_P(mu,P)
% Generate a single random sample from a Gaussian PDF with mean vector mu
% and covariance matrix P. The mean must be a column vector if dim > 1.
n=size(mu,1);
if length(mu)~=n
    error('Mean must be a column vector')
end
if (n~=size(P,1))||(n~=size(P,2))
    error('Dimension of mu not compatible with P')
end
P=0.5*(P+P'); % ensure symmetry
T=transpose(chol(P)); % Chloesky factorisation of cov matrix such that P = T*T'
g=randn(size(mu)); % Generate random sample from N(0, I)
g = T*g + mu; % Affine transformation to N(mu,P)
return
\end{verbatim}

\subsection{Confusion Matrix code}\label{confus}

\begin{verbatim}
function [C,TP,FP,FN,TN,PA,MIoU] = confusion_matrix(truth,pred,num_class)
% Compute confusion matrix C & performance metrics for multi-class classifier
% with num_class classes, given vector of true class indexes,
% and vector of predicted classes from classifier obtained on test data.
% For each class, compute the class-conditioned TP, FP, FN and TN.
% The overall accuracy and mean intersection over union (MIoU) are also computed.
% An error results if a truth or pred value is found that exceeds num_class-1
% since the function assumes that values are in [0, 1, ..., num_class-1].
%
% Copyright 2020 Graham Pulford

N1=length(truth); % number of truth samples
N2=length(pred); % number of predicted samples
if N1~=N2
    error('Number of predicted samples differs from number of truth samples')
end
truth_values=unique(truth);
if min(truth_values)<0 || max(truth_values)>num_class-1
    error('truth values must be in the range [0:num_class-1]')
end
pred_values=unique(pred);
if min(pred_values)<0 || max(pred_values)>num_class-1
    error('predicted values must be in the range [0:num_class-1]')
end

C=zeros(num_class,num_class);
for n=1:N2
    i=truth(n);
    j=pred(n);
    C(i+1,j+1)=C(i+1,j+1)+1;
end

TP=zeros(1,num_class); % True positives
FP=zeros(1,num_class); % False positives
FN=zeros(1,num_class); % False negatives
TN=zeros(1,num_class); % True negatives
for i=1:num_class
    TP(i)=C(i,i);
    FP(i)=sum(C(:,i))-C(i,i);
    FN(i)=sum(C(i,:))-C(i,i);
    TN(i)=sum(sum(C))-TP(i)-FP(i)-FN(i);
end

% Overall ccuracy = sum of class-conditioned accuracies
PA=sum(diag(C))/sum(sum(C));
% Mean intersection over union = average Jaccard coefficient
MIoU=0;
for i=1:num_class
    MIoU=MIoU+C(i,i)/(sum(C(i,:))+sum(C(:,i))-C(i,i));
end
MIoU=MIoU/num_class;
return
\end{verbatim}

\end{document}